\documentclass[10pt,twocolumn,letterpaper]{article}

\usepackage[pagenumbers]{wacv} %

\usepackage{times}
\usepackage{epsfig}
\usepackage{graphicx}
\usepackage{amsmath}
\usepackage{amssymb}
\usepackage{booktabs}
\usepackage[table,xcdraw]{xcolor}
\usepackage{pifont}
\usepackage{array}
\usepackage[ruled]{algorithm2e}
\usepackage{diagbox}
\newcolumntype{H}{>{\setbox0=\hbox\bgroup}c<{\egroup}@{}}
\newcommand{\cmark}{\faCheck}%
\newcommand{\cmarkcircle}{\faCheckCircle}%
\newcommand{\xmark}{\faTimes}

\definecolor{Gray}{gray}{0.95}
\definecolor{highlight}{rgb}{0.824,0.976,0.824}
\definecolor{darkgray}{rgb}{0.902,0.898,0.902} 
\definecolor{dim}{rgb}{0.95,0.95,0.95}
\definecolor{uniform}{rgb}{0.702,0.702,0.702}
\definecolor{delta-color}{rgb}{0.824,0.824,0.976}
\definecolor{baseline}{gray}{0.9}
\usepackage{colortbl}
\usepackage[font=small,labelfont=bf]{caption}
\usepackage{enumitem}
\usepackage{multirow}
\usepackage{multicol}
\usepackage{fontawesome}
\usepackage{colortbl}

\usepackage[pagebackref,breaklinks,colorlinks]{hyperref}

\usepackage[capitalize]{cleveref}
\crefname{section}{Sec.}{Secs.}
\Crefname{section}{Section}{Sections}
\Crefname{table}{Table}{Tables}
\crefname{table}{Tab.}{Tabs.}

\newcommand{\smethodnames}{$\Delta$-modules\xspace}
\newcommand{\smethodname}{$\Delta$-module\xspace}
\newcommand{\settingname}{$\Delta$-Patching\xspace}
\newcommand{\methodname}{$\Delta$-Networks\xspace}

\newcommand{\ve}[1]{\mathbf{#1}} %
\newcommand{\ma}[1]{\mathrm{#1}} %

\def\wacvPaperID{1898} %
\def\confName{WACV}
\def\confYear{2024}

\begin{document}

\title{\settingname: A Framework for Rapid Adaptation of Pre-trained Convolutional Networks without Base Performance Loss}

\author{Chaitanya Devaguptapu\textsuperscript{1}\thanks{Major part of the work was done when Chaitanya was a Visiting Graduate Researcher at University of Toronto and a Masters student at IIT-Hyderabad. Corresponding author: email@chaitanya.one} \hspace{0.5cm} Samarth Sinha\textsuperscript{2,4} \hspace{0.5cm}  K J Joseph\textsuperscript{6}\\
Vineeth N Balasubramanian\textsuperscript{3} \hspace{1cm} Animesh Garg\textsuperscript{2,4,5} \\\\
\textsuperscript{1}Fujitsu Research India \hspace{0.25cm}\textsuperscript{2}University of Toronto  \hspace{0.25cm} \textsuperscript{3}Indian Institute of Technology, Hyderabad\hspace{0.25cm} \\\textsuperscript{4}Vector Institute\hspace{0.25cm} \textsuperscript{5}NVIDIA\hspace{0.25cm} \textsuperscript{6}Adobe Research\\
}
\maketitle

\begin{abstract}
Models pre-trained on large-scale datasets are often fine-tuned to support newer tasks and datasets that arrive over time. This process necessitates storing copies of the model over time for each task that the pre-trained model is fine-tuned to. Building on top of recent model patching work, we propose $\Delta$-Patching for fine-tuning neural network models in an efficient manner, without the need to store model copies. We propose a simple and lightweight method called $\Delta$-Networks to achieve this objective. Our comprehensive experiments across setting and architecture variants show that $\Delta$-Networks outperform earlier model patching work while only requiring a fraction of parameters to be trained. We also show that this approach can be used for other problem settings such as transfer learning and zero-shot domain adaptation, as well as other tasks such as detection and segmentation. 
\end{abstract}

\section{Introduction}
\label{sec_intro}

\noindent The proliferation of pre-trained models, especially in the domain of vision tasks, has been remarkable~\cite{hugging-face-2022}. These models, trained on extensive datasets, possess features that are universally beneficial across a wide array of tasks. Over the years, a significant use of such pre-trained models has been in a fine-tuning or transfer learning context, where the model weights are partially or fully modified while maximizing performance on a target task. While effective, this strategy tends to specialize the model's features, often at the expense of its original capabilities, causing a decline in performance on the original task for which the model was initially trained. Recently, there have been efforts to ``\textit{patch}'' pre-trained models  towards improved performance on different target tasks with minimal loss of performance on the original task \cite{modelpatching2022}. 
\textit{Model patching}~\cite{modelpatching2022} draws parallels with software patching in software engineering and aligns with the growing trend of treating machine learning models as open-source software~\cite{ml-like-software-2021, modelpatching2022}.

\begin{figure}
  \includegraphics[width=0.96\linewidth]{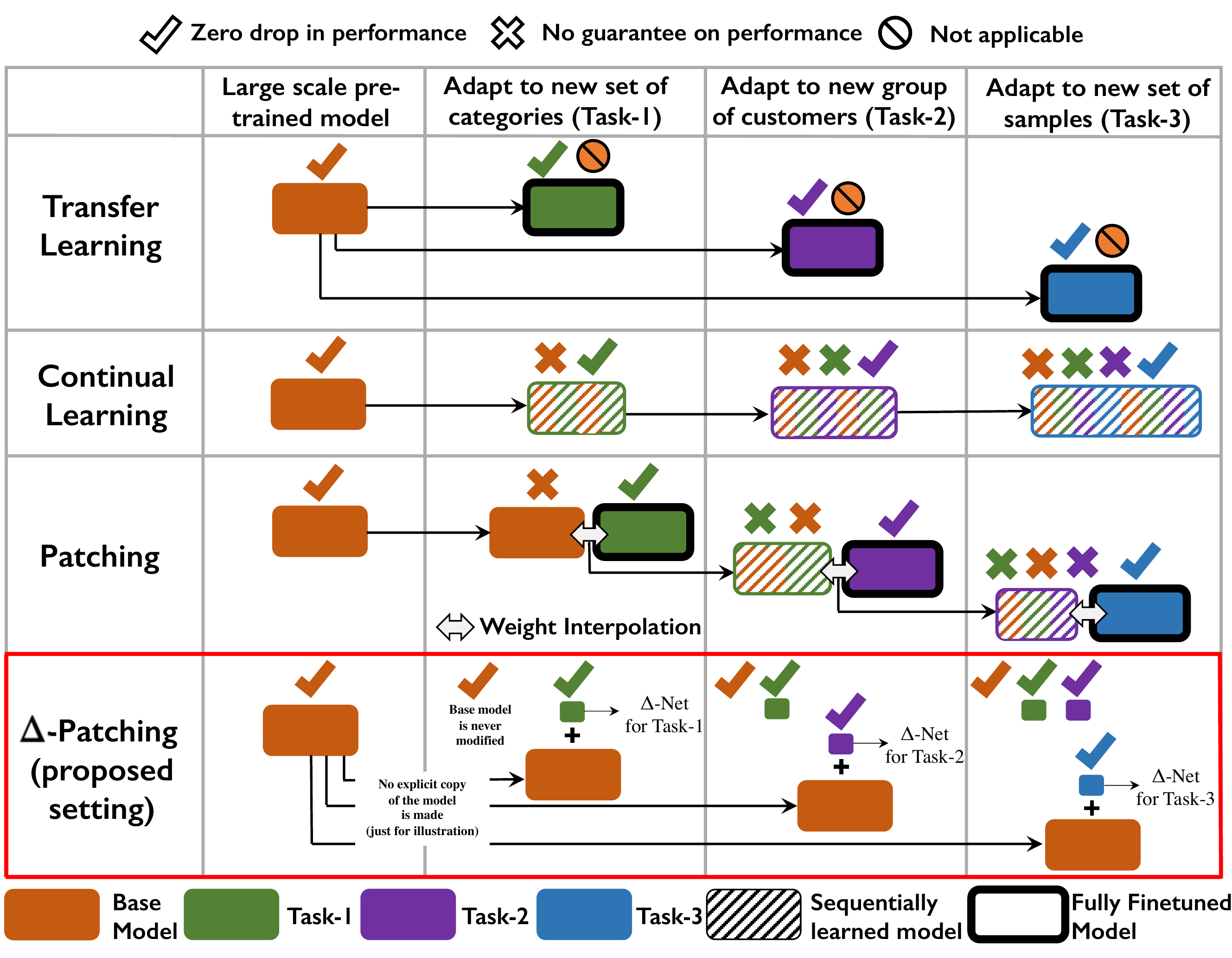}
  \caption{Comparison of \settingname with relevant learning with limited data paradigms. \settingname ensures there is \textbf{\textit{zero-drop in performance}} on the base task and \methodname (proposed approach to perform \settingname) use \textbf{\textit{significantly less trainable parameters}} as compared to existing approaches} 
  \label{fig:teaser}
  \vspace{-0.55cm}
\end{figure}

\noindent In this work, we propose \settingname a novel framework engineered for the swift adaptation of large-scale pre-trained models to new tasks \textit{without} degrading accuracy on existing tasks. The impetus for \settingname arises from practical needs observed in various industries.  Consider, for example, an e-commerce platform equipped with a large-scale apparel classification model. As fashion trends evolve or seasons change, the model pre-trained on a comprehensive dataset encompassing a multitude of apparel categories, must adapt to evolving fashion trends and seasonal variations without losing its proficiency in recognizing existing categories. In the medical realm, a healthcare system employing a pre-trained model for medical image diagnosis might need to detect a newly identified condition from MRI scans, without diminishing its diagnostic accuracy for established conditions like tumors. Contrasting with traditional Incremental/Continual Learning (IL) methods, which risk "catastrophic forgetting" due to continual modification of the base model, \settingname adopts a different philosophy. It capitalizes on the same large-scale pre-trained model for all tasks, ensuring the preservation of its intrinsic knowledge. While IL methods grapple with balancing new task adaptation and old task performance retention, \settingname  bypasses this dilemma. It promises rapid and adaptable integration of new tasks while safeguarding the original capabilities of the pre-trained model. This positions \settingname as a compelling strategy, especially when rapid, flexible adaptation to new tasks, without compromising existing capabilities, is paramount.

\noindent As an embodiment to achieve \settingname, we introduce \methodname.  While not entirely novel, \methodname excel in terms of efficiency and ease of adaptability, making it a highly effective solution for \settingname. \methodname is a hypernetwork~\cite{hypernetwork2016} with a twist: it incorporates input-conditioned skip connections for adaptive and efficient patching, all while not modifying a single weight of the original model. 
While sharing the overarching goal of model adaptability with existing work~\cite{modelpatching2022}, our approach distinguishes itself in several key aspects. First, while \cite{modelpatching2022} focuses on patching open-vocabulary models like CLIP \cite{pmlr-v139-radford21a}, we focus on general-purpose CNNs, which are ubiquitously used and fine-tuned across various domains. Unlike PAINT~\cite{modelpatching2022}  that necessitate dual model copies for interpolation \cite{modelpatching2022}, \methodname introduces input-conditioned skip connections for patching, leaving the original model weights untouched. The essence of our approach, symbolized by the $\Delta$ in $\Delta$-patching and $\Delta$-networks, lies in the minimal input-conditioned skip connections added. This design facilitates memory-efficient model patching, adaptable to varying memory and computational constraints. Crucially, \methodname is fully differentiable, eliminating the need for model interpolations or empirically determined weight coefficients.  Figure \ref{fig:teaser} illustrates the proposed setting and its difference from earlier work. Table \ref{tab:comparison} presents a feature-wise comparison of the setting when compared to similar settings in related work. 
\noindent While \methodname are primarily engineered for \settingname, they support a wide range of tasks, like object detection, segmentation and can also be used to improve zero-shot domain adaption, transfer learning. The simplicity of our method is its strength. In fact, \methodname achieve new state-of-the-art for transfer learning by improving the performance of an existing transfer learning method. Our key contributions can be summarized as follows:
\vspace{-5pt}
\begin{itemize}[leftmargin=*]
\setlength\itemsep{-0.2em}
    \item Introduction of \settingname, a new model patching strategy that enables rapid adaptation of pre-trained models to new tasks without sacrificing original performance, implemented through \methodname.
    \item Development of an end-to-end differentiable method that allows patching at varying levels based on computational constraints.
    \item Extensive evaluation of our methodology across multiple datasets and tasks, demonstrating superior performance with fewer trainable parameters, and setting a new state-of-the-art in transfer learning.
\end{itemize}

\section{\methodname for Efficient Model Patching}
\label{sec_method}
\vspace{-3pt}
\noindent Consider a model $\mathcal{M_S}$ trained for a base task $\mathcal{T_S}$ using samples from a dataset $\mathcal{D_S}$. Our primary goal is to \textit{patch} $\mathcal{M_S}$ for performing well on a new task $\mathcal{T_P}$ with samples from  $\mathcal{D_P}$ without any performance degradation on $\mathcal{T_S}$. Importantly, we operate under the practical constraints that neither the samples $\mathcal{D_S}$ nor any samples generated using $\mathcal{M_S}$ are accessible during this process. Let the features learned using $\mathcal{D_S}$ for the base/supporting task $\mathcal{T_S}$ be denoted as $\mathcal{F_S}$.  $\mathcal{F_S}^k$ then denotes a layer (or a block of layers) (for e.g. with convolutional, max-pooling operations and activation functions in a traditional CNN). $\mathcal{F_S}$, the final representation learned with samples from $\mathcal{D_S}$ can be viewed as a composition of features learned at each layer, and can be written as below for a given input sample $\mathbf{x}_s$ from the dataset $\mathcal{D_S}$.
\vspace{-4pt}
\begin{equation}
    \mathcal{F_S} = \mathcal{F_S}^k(\mathcal{F_S}^{k-1} (\dots (\mathcal{F_S}^1(\mathbf{x_s})))) 
    \label{eq:1}
    \vspace{-2pt}
\end{equation}
\noindent The final model $\mathcal{M_S}$ is formed by passing the features $\mathcal{F_S}$ into a classification layer/network, denoted using $\phi_{S}$ which typically is a fully-connected layer with neurons equal to the number of classes in $\mathcal{D_S}$:
\vspace{-4pt}
\begin{equation}
    \mathcal{M_S} = \mathcal{F_S} \circ \phi_{S}  
    \label{eq:2}
    \vspace{-4pt}
\end{equation}
\noindent When a new patching task $\mathcal{T_P}$ arrives with a corresponding dataset $\mathcal{D_P}$, the number of classes in the patching task is generally different from that of the original task. So, adding a new classification layer, $\phi_{P}$ is most often inevitable. Directly modifying $\mathcal{F_S}$ to perform better on $\mathcal{T_P}$ affects the performance on $\mathcal{T_S}$, which defeats our primary goal. We hence introduce an architectural modification to $\mathcal{M_S}$ in such a way that it helps address the new task while not affecting the performance on the original task. 

\noindent As mentioned previously, $\mathcal{F}^k$ denotes a sub-block of layers inside $\mathcal{M_S}$. Most modern-day architectures stack sub-blocks to build blocks which are again stacked together to build larger architectures. In a general-purpose CNN architecture such as ResNet \cite{resnets2015}, $\mathcal{F}^k$ is a residual block with one skip connection (we refer to this as a sub-block). %
The output representation for any $\mathcal{F}^j$ which represents a sub-block of layers with one skip connection can be denoted as follows:
\vspace{-3pt}
\begin{equation}
    \mathcal{F}^j = \mathcal{F}^j(\mathcal{F}^{j-1} (\dots (\mathcal{F}^1(\mathbf{x}))); \mathbf{W}_j) + \underbrace{{\textcolor{red}{\mathcal{F}^{j-1}(\mathbf{x}; {S_w})}}}_{\text{Skip connection}} 
    \label{eq:3}
    \vspace{-4pt}
\end{equation}
\noindent where $\mathbf{W_j}$ denotes the weights corresponding to the $j^{\text{th}}$ sub-block and the highlighted term in Eqn \ref{eq:3} refers to the representation learned by the previous sub-block, mapped using a weight matrix ${S_w}$. In a standard residual network, this highlighted component is obtained by setting ${S_w} = I$, where $I$ is the identity matrix. Generalizing the notion of skip connections using ${S_w}$ is a key component of the proposed \methodname methodology. 

\begin{figure}[h]
 \centering \includegraphics[width=0.7\linewidth]{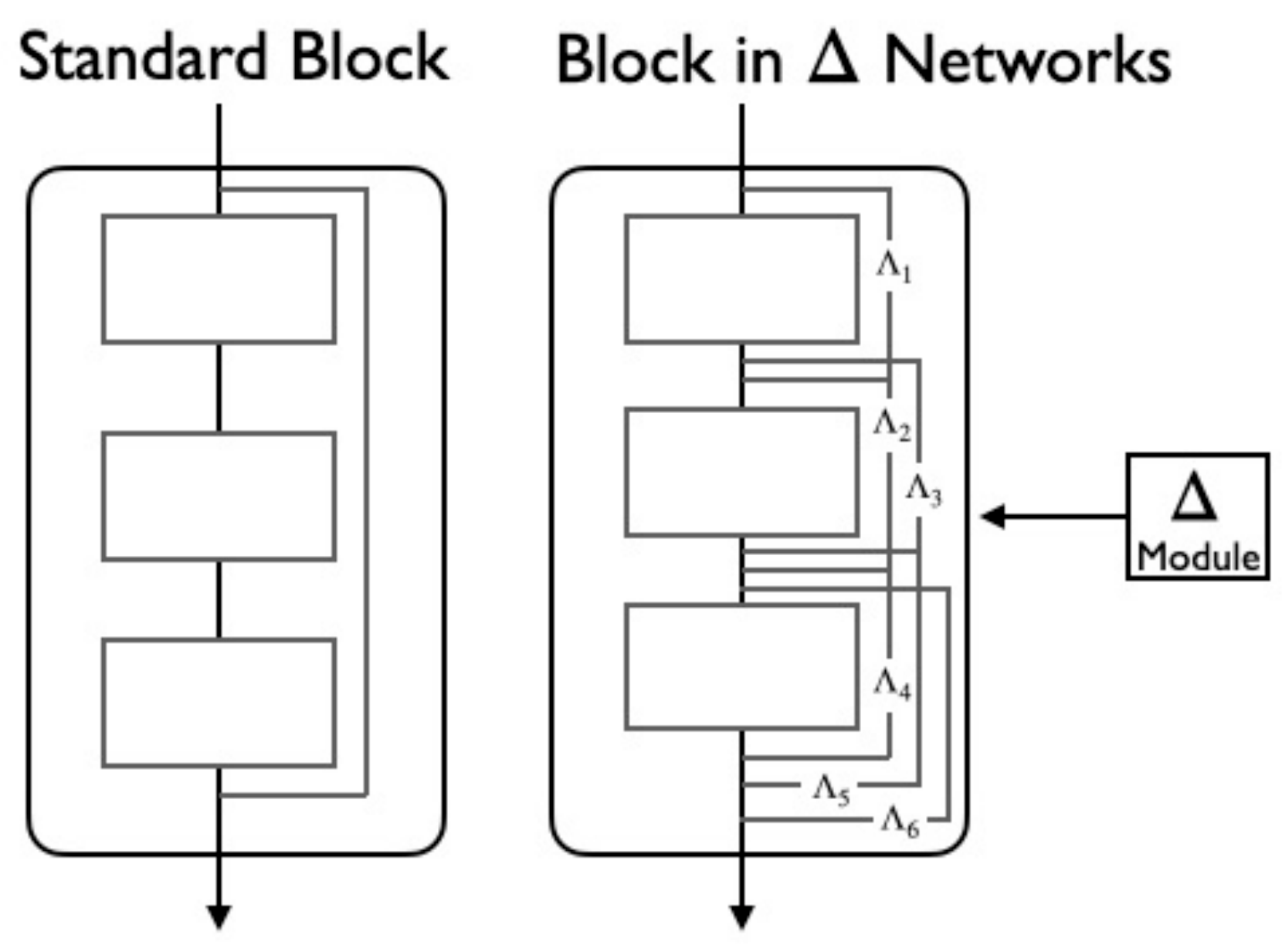}
  \vspace{-8pt}
  \caption{Illustration of a block in \methodname} 
  \label{fig:method-figure}
  \vspace{-0.4cm}
\end{figure}

\noindent While most architectures as well as earlier transfer learning/patching literature focus on updating $\mathbf{W}_j$, we instead turn our focus to the above generalization of skip connections to leverage the features learned on the base/original task. Firstly, we consider all possible skip connections between each sub-block in a block as possible routes for a given input. (Note that we do not introduce skip connections inside layers of a sub-block to keep the overhead minimal.) Figure \ref{fig:method-figure} shows an illustration.

\noindent One could view this as similar to a sparse variant of DenseNets, however with a difference -- each connection has a learnable weight matrix ${S_w}$ in our case, as described later in this section. The output of a block in Eqn \ref{eq:3} can hence be written as: 
\vspace{-6pt}
\begin{equation}
    \mathcal{F}^j = \mathcal{F}^j(\mathcal{F}^{j-1} (\dots (\mathcal{F}^1(\mathbf{x}))); \mathbf{W}_j) + {\textcolor{red}{\sum_{i=1}^{j-1}\mathcal{F}^{i}(\mathbf{x}; {S_w})}} 
    \label{eq:4}
    \vspace{-4pt}
\end{equation}
\noindent where ${S_w}$ is learnable and specifies how two sub-blocks interact. 
Since this modification does not involve any changes to existing model weights, off-the-shelf pre-trained models can be easily leveraged. Besides, not making any changes to the base model ensures that the performance on the base task is retained without any drop (by simply dropping the added skip connections). Through this generalization, our objective can be viewed as developing a mechanism that allows us to control interactions among pre-trained features at various granularities (for e.g, early vs late layers) so that it enables effective patching for new tasks.

\vspace{3pt}
\noindent \textbf{The \smethodname.} When learning ${S_w}$s for skip connections on top of a pre-trained model for a target task, the same weights that are optimal for one input might be sub-optimal for another input, even from the same dataset or class label. Keeping this in mind, we propose the use of a \smethodname which is a separate, small and lightweight network parameterized by weights $\theta$ that learns the weights of the additional skip connections independent of the base model.
Thus, for a given input $\mathbf{x}$, the added skip connections are weighted in an input-conditioned manner using a \smethodname. Following Eqn \ref{eq:4}, the output of a sub-block $\mathcal{F}_j$ is now changed as follows when a \smethodname (denoted using $\Delta$ below) is introduced: 
\vspace{-4pt}
\begin{equation*}
     \textcolor{red}{\Lambda_{i} = \Delta(\mathbf{x}; \theta)}
    \vspace{-9pt}
\end{equation*}
 \begin{equation*}
    \mathcal{F}^j (\textbf{x})= \mathcal{F}^j(\mathcal{F}^{j-1} (\dots (\mathcal{F}^1(\mathbf{x}))); \mathbf{W}_j) + {\sum_{i=1}^{j-1}\mathcal{F}^{i}(\mathbf{x}; \textcolor{red}{\Lambda_i})} 
 \vspace{-3pt}
 \end{equation*}

\noindent Intuitively, we first pass the input to the \smethodname to obtain weighting coefficients $\Lambda_{\{1, \dots (j - 1)\}}$ for each of the new skip connections in our $\Delta$-network. 
Since the \smethodname is a parametrized neural network and therefore fully differentiable, it can be trained directly on the new task loss via gradient descent.
Unlike standard skip connections with $\Lambda_i = I$, adaptive skip connections have input-conditioned weights on them instead. 
Since the skip connections are introduced after training, our model can be directly used with existing models that are pre-trained for a different task. (For models that already have skip connections -- for example, DenseNets, the $\Lambda_i$s for those pre-existing skip connections are defaulted to the identity matrix when addressing the base task.) 
An end-to-end summary of our approach is provided in Algorithm \ref{alg:end-to-end}.

\vspace{-5pt}
\begin{algorithm}[h]
\footnotesize
\SetAlgoLined
\textbf{Inputs and Initialization:} \\ 
\begin{itemize}[noitemsep,leftmargin=*]
    \item Model trained on supporting task $\mathcal{M_S} = \mathcal{F_S} \circ \phi_{S}$, $\mathcal{F_S} \rightarrow \mathcal{F_S}^{1 \dots k}$ with $k$ blocks; \item Randomly initialized classification head $\phi_{P}$, and \methodname, $\Delta^{1 \dots k}$
    \item Samples from patching task $\mathbf{x_d}^{m=1 \dots n} \in \mathcal{D_P}$ in $n$ batches, Loss function used for the task $\mathcal{L}$ 
    \item Add $\phi_T$ to $\mathcal{F_S}$; Introduce all possible skip connections ($p$) in each sub-block of $\mathcal{F_S}^{1 \dots k}$
\end{itemize}
\textbf{Training:} \\ 
\For{$\text{num\_epochs}$}{
    \For{$\mathbf{x_d}^{m=1 \dots n}$}{
        \For{$j = 1 \dots k$}{
            $\Lambda_{1 \dots p} = \Delta^j(\mathbf{x_d^m}; \theta_j)$ \\
            ${\displaystyle \mathcal{F_S}^j = \mathcal{F_S}^j(\mathbf{x_d^m}; \mathbf{W_j}) + \sum_{i=1}^{j-1} \mathcal{F_S}^i(\mathbf{x_d^m}; \Lambda_i) }$ \\
        }
        $\mathcal{M_P} =  (\mathcal{F_S}, \Delta^{1 \dots k}) \circ \phi_{T}$ \\ 
        Calculate $\mathcal{L}$  \\ 
        Update $\Delta^{k}$, $\phi_T$  \\
    }
}
\textbf{Output:} $ \mathcal{F_S} \circ \phi_{T} \rightarrow$ model patched for task $\mathcal{T_P}$  \\
\caption{\methodname for Model Patching}
\label{alg:end-to-end}
\end{algorithm}
\vspace{-5pt}

\noindent Conditioning skip connections using an external fully-differentiable network 
helps finetune the model easily on a new patching task. When new tasks arrive in a sequence, we introduce a new \smethodname and a classification layer for each task. These \smethodnames are intended to be lightweight (2-layer feedforward networks in our implementation), and hence can be stored easily for each task arriving over time. %
As each task has its own classifier and \methodname, there is no loss of performance on any of the preceding tasks and the model can support as many tasks as required. %

\noindent \textbf{Adapting to Memory/Compute Requirements.} When using Algorithm \ref{alg:end-to-end},  the memory and compute requirements can be controlled using %
the number of \smethodnames used for a task under consideration. When there are memory/compute constraints, one can choose to use a small \smethodname for learning the weights of skip connections of a few blocks alone. $\Delta$ hence becomes a unit quantum of additional overhead, lending the name \methodname to this approach. With more memory, a set of \smethodnames can be used to learn different skip connections on the base model. The proposed algorithm thus allows customizing \settingname in a memory-sensitive manner. Increasing the number of \smethodnames increases the number of trainable parameters, and hence results in performance gain in general. We provide some design choices and thumb rules of connecting \smethodname to subsets of skip connections in the following section

\noindent \textbf{Implementation and Design Details.}
Any neural network model developed by stacking sub-blocks of layers can be segregated into a set of blocks, denoted as $\mathcal{F}^j$ earlier in this section, with each block containing varying number of sub-blocks. For example, ResNet architectures are often divided into four blocks.
If a block contains $k$ sub-blocks, we introduce a total $k(k+1)/2$ skip connections in the block, which is controlled by one \smethodname.
One can use additional \smethodnames to weight skip connections in other blocks. We denote \methodname $(n)$ to denote a $\Delta$-network with $n$ \smethodnames.
Considering that we only need to learn the weights of the skip connections (and not all parameters of the network), each of our \smethodname consists of a simple two-layer CNN with ReLU activations, normalization layers, and a fully-connected layer to determine the weights for skip connections. The output of a \smethodname is constrained between $[0,1]$ using a sigmoid activation. Whenever skip connections are present in the architecture, there is a problem of increase in variance~\cite{fixupinit} (we formally show this in the Appendix). With \methodname, when all possible skip connections are introduced, the variance increases rapidly since we add features from all preceding blocks. We address this issue using one BN layer in each \smethodname in our implementation.

\section{Experiments and Results}
\noindent In this section, we aim to provide a holistic evaluation of \methodname, contextualizing its performance against both established and alternative baselines. We first compare \methodname with other relevant methods adapted for the patching setting, and then proceed to a detailed empirical analysis against PAINT\cite{modelpatching2022}, the established baseline for model patching.

\noindent Beyond patching, we also explore the applicability of 
\methodname to a range of other tasks, such as object detection and segmentation. Additionally, we examine its performance in specialized settings like zero-shot domain adaptation and transfer learning. To delve deeper into the underlying factors contributing to the success of \methodname, we conclude this section with an ablation analysis.

\noindent \textbf{Experimental Setup:} \noindent For all the experiments, we leverage CNN models pre-trained on ImageNet, sourced from the official PyTorch \cite{paszke2019pytorch} and PyTorch Image Models \cite{rw2019timm} libraries. To ensure a fair comparison, we employ the official code repository of PAINT\cite{modelpatching2022} and apply their method to all models under consideration in our study. In all experiments, both for \methodname and PAINT, we append a fully connected (FC) layer to the model architecture for the new task, consistent with the approach in PAINT. Since PAINT mandates fine-tuning of the entire set of pre-trained weights, we adhere to a 50-epoch fine-tuning regimen for all comparative experiments, utilizing the same data splits and procedures as described in PAINT, to maintain fairness in evaluation. For the experiments where we compare with adapted baselines, we follow the data splits and use the code provided in their official code repositories\cite{Rebuffi17, rebuffi-cvpr2018,mirza2022norm}. 
Our primary performance metric is the harmonic mean of top-1 accuracies for both pre-trained and fine-tuned tasks, except in the case of sequential patching where we use the mean. Detailed top-1 accuracy scores for individual tasks are available in the appendix for a comprehensive understanding.
\noindent When deploying \methodname, we introduce one, two, or four \smethodnames. A single \smethodname is added to the last block of the base architecture, complete with corresponding skip connections. For configurations with two or four \smethodnames, they are distributed across the last two or all four blocks of the network, respectively. Each  \smethodname incrementally increases the number of trainable parameters, offering a lens through which to study the scalability of \methodname with respect to parameter size and hardware constraints.

\subsection{Evaluation of \methodname for Patching} \label{sec:patching}
\vspace{-4pt}

\begin{table*}[h]
\resizebox{\linewidth}{!}{\begin{tabular}{>{\kern-\tabcolsep}l|ccccccccc<{\kern-\tabcolsep}}
\toprule
                              & Flowers        & Aircraft       & UCF101         & GTSRB          & DTD            & SVHN           & CIFAR-100      & OmniGlot       & Daimler Ped    \\ \midrule
PAINT                         & 12.73          & 4.94           & 21.33          & 21.36          & 8.54           & 62.49          & 9.27           & 9.22           & 18.58          \\ \midrule
Parallel Adapters \cite{Rebuffi17}             & 71.46          & 52.01          & 73.21          & 83.20          & 55.68          & 81.30          & 75.52          & 77.59          & 83.15          \\
Series Adapters \cite{rebuffi-cvpr2018}              & 59.10          & 46.65          & 70.67          & 83.20          & 39.97          & 81.34          & 70.40          & 77.94          & 83.11          \\ \midrule
BatchNormAdapt \cite{mirza2022norm} & 71.52          & 52.52          & 73.23          & 83.00          & 53.20          & 81.28          & \textbf{75.70} & 78.07          & 83.07          \\
\rowcolor{highlight} \methodname (4)               & \textbf{75.16} & \textbf{58.44} & \textbf{75.52} & \textbf{83.21} & \textbf{56.06} & \textbf{81.52} & 75.12          & \textbf{78.40} & \textbf{83.16} \\ \bottomrule
\end{tabular}}
\vspace{-7pt}
\caption{\textit{\textbf{\methodname excels in 8/9 datasets.}} Comparative Analysis vs. Adapted Methods in \settingname context. Harmonic Mean of Top-1 Accuracies reported.}
\label{tab:related-approaches}
\vspace{-18pt}
\end{table*}

\begin{table}[h]
\resizebox{\linewidth}{!}{\begin{tabular}{@{}l|l|l|c|c|c|c|c|c@{}}
\toprule
Architecture                   & \multicolumn{1}{c|}{Method}    & \# Parameters & STL10          & RESISC45       & GTSRB          & Flowers        & Cars           & Aircraft       \\ \midrule
\multirow{4}{*}{ResNet-18}     & PAINT                          & 11.2M         & 55.41          & 24.38          & 18.39          & 12.70          & 2.01           & 3.51           \\
                               & \methodname (1) & 0.4M          & 79.55          & 74.48          & 70.22          & 75.65          & 46.78          & 46.56          \\
                               & \methodname (2) & 0.6M          & \textbf{79.66} & 76.08          & 76.20          & \textbf{75.91} & 51.29          & 50.48          \\
                               & \methodname (4) & 0.6M          & 79.57          & \textbf{76.42} & \textbf{78.94} & 75.65          & \textbf{76.23} & \textbf{52.67} \\ \midrule
\multirow{4}{*}{ResNet-34}     & PAINT                          & 21.8M         & 62.64          & 21.91          & 19.87          & 16.16          & 4.39           & 2.94           \\
                               & \methodname (1) & 0.4M          & 83.41          & 78.02          & 70.83          & 78.31          & 47.31          & 45.74          \\
                               & \methodname (2) & 0.6M          & 83.59          & 80.18          & 78.11          & \textbf{79.04} & 55.24          & 51.39          \\
                               & \methodname (4) & 0.6M          & \textbf{83.60} & \textbf{80.70} & \textbf{82.36} & 78.82          & \textbf{57.33} & \textbf{54.82} \\ \midrule
\multirow{4}{*}{ResNet-50}     & PAINT                          & 23.5M         & 86.30          & 48.04          & 28.11          & 56.97          & 12.54          & 7.33           \\
                               & \methodname (1) & 7.1M          & \textbf{87.53} & 84.02          & 77.79          & 82.64          & 59.31          & 55.81          \\
                               & \methodname (2) & 8.9M          & 87.41          & 84.93          & 84.09          & \textbf{82.74} & 65.96          & \textbf{62.14} \\
                               & \methodname (4) & 9.3M          & 87.40          & \textbf{85.03} & \textbf{85.67} & 80.34          & \textbf{67.11} & 61.48          \\ \midrule
\multirow{4}{*}{ConvNeXT-Tiny} & PAINT                          & 27.8M         & 84.57          & 34.81          & 12.36          & 69.06          & 6.87           & 6.18           \\
                               & \methodname (1) & 4.0M          & \textbf{89.69} & 86.54          & 82.47          & \textbf{87.49} & 73.00          & 69.59          \\
                               & \methodname (2) & 5.0M          & 89.32          & 86.97          & 88.01          & 84.89          & 75.44          & \textbf{71.74} \\
                               & \methodname (4) & 5.3M          & 89.35          & \textbf{87.24} & \textbf{88.96} & 82.35          & \textbf{76.23} & 70.81          \\ \bottomrule
\end{tabular}}
\vspace{-7pt}
\caption{\textit{\textbf{\methodname outperforms PAINT with fewer parameters.}} Comparison in \textit{\textbf{Single-Task Patching}} across architectures. Harmonic Mean of Top-1 accuracies reported.}
\label{tab:patching-one-task}
\vspace{-20pt}
\end{table}

\noindent \textbf{Comparison with Adapted Baselines}: To situate \methodname within the broader landscape of model patching, we adapt existing methods such as Parallel and Series Adaptors\cite{Rebuffi17, rebuffi-cvpr2018}, as well as a recent normalization-based approach\cite{mirza2022norm}, to the patching setting. The harmonic mean of Top-1 accuracies serves as our performance metric, as mentioned earlier. As shown in Table \ref{tab:related-approaches}, \methodname outperforms these adapted methods on 8 out of 9 datasets, thereby establishing its efficacy.

\noindent An additional baseline that warrants consideration in the context of patching is fine-tuning based solely on a Fully Connected (FC) layer. This approach serves as a fundamental comparison point, given its widespread use and simplicity. We delve into this comparison in greater detail later in the paper, presenting the results in Table \ref{tab:ablation}.

\noindent \textbf{Evaluating \methodname in various patching scenarios}: \noindent In alignment with the experimental setup outlined in PAINT \cite{modelpatching2022}, we evaluate \methodname across three distinct patching scenarios: (1) \textit{Single-Task Patching}, where the model is adapted for one new task; (2) \textit{Joint Patching}, involving adaptation for a pre-defined set of tasks; and (3) \textit{Sequential Patching}, where the model is adapted for tasks arriving in sequence. 

\noindent \textbf{Single-Task Patching:}
Table \ref{tab:patching-one-task} presents the outcomes of our experiments in a single-task setting. We explore six diverse datasets and four architectural variations. Additional results for other architectural variants can be found in the Appendix. For rows related to PAINT, we select the model interpolation coefficient that maximizes the harmonic mean metric previously discussed. \noindent As evident from the table, \methodname consistently outperforms PAINT by substantial margins while utilizing a significantly lower number of trainable parameters.
As we only finetune the fully-connected layer and the small \smethodname, the number of trainable parameters are much fewer compared to full fine-tuning approaches like PAINT. Increasing the number of \smethodnames by introducing them in the first, second and third blocks along with the last block increases the performance even more.

\begin{table}[h]
\resizebox{\linewidth}{!}{\begin{tabular}{@{}l|l|c|c|c@{}}
\toprule
\textbf{Architecture} & \textbf{Method}    & \multicolumn{1}{l|}{\textbf{\begin{tabular}[c]{@{}l@{}} Aircraft\\Flowers\\STL10\end{tabular}}} & \multicolumn{1}{l|}{\textbf{\begin{tabular}[c]{@{}l@{}} GTSRB\\Cars\\Flowers\end{tabular}}} & \multicolumn{1}{l}{\textbf{\begin{tabular}[c]{@{}l@{}} STL10\\Cars\\GTSRB\end{tabular}}} \\ \midrule

ResNet-18     & PAINT              & 22.33                           & 4.35                        & 0.66                      \\
              & \methodname (1) & 66.91                           & 61.41                       & 66.91                     \\ 
             & \methodname (2) & 69.04                           & 64.71                       & 69.45                     \\
              & \methodname (4) & \textbf{71.14}                           & \textbf{66.64}                       & \textbf{71.14}                     \\ 
             \midrule
ResNet-34     & PAINT              & 13.75                           & 6.75                        & 15.20                     \\
              & \methodname (1) & 70.31                           & 64.51                       & 70.31                     \\ 
              & \methodname (2) & \textbf{82.48}                           & 67.81                       & 72.93                     \\
              & \methodname (4) & 75.39                           & \textbf{71.32}                       & \textbf{75.39}                     \\
              \midrule
ResNet-50     & PAINT              & 55.65                           & 4.70                        & 35.16                     \\
              & \methodname (1) & 76.56                           & 71.96                       & 76.56                     \\ 
              & \methodname (2) & 80.61                           & 76.94                       & 80.02                     \\
              & \methodname (4) & \textbf{81.79}                           & \textbf{78.75}                       & \textbf{81.79}                     \\
              \midrule

ConvNeXt-TINY & PAINT              & 34.72                           & 1.61     & 6.70                      \\
              & \methodname (1) & 81.12                           & 77.45                       & 81.02                     \\    
              & \methodname (2) & 82.69                           & 80.91                       & 83.29                     \\
              & \methodname (4) & \textbf{82.91}                           & \textbf{82.32}                       & \textbf{84.33}                     \\ 
              \bottomrule
\end{tabular}}
\vspace{-10pt}
\caption{Performance Evaluation of \methodname and PAINT\cite{modelpatching2022} in \textbf{\textit{Joint Patching}} scenario. Harmonic Mean of Top-1 accuracies is reported.}
\label{tab:joint-patching}
\vspace{-10pt}
\end{table}

\noindent \textbf{Joint Patching:} In this experimental setting, we have access to all the datasets for the tasks that need to be learned simultaneously. In alignment with \cite{modelpatching2022}, we concatenate the training and test sets for all tasks prior to fine-tuning. The newly introduced fully connected layer comprises neurons equal to the total number of classes across all tasks. \smethodnames are trained in a unified manner for all tasks. Table \ref{tab:joint-patching} displays the results, which demonstrate that our method consistently outperforms PAINT across all task and architecture variants. 

\begin{table}[]
\centering
\resizebox{\linewidth}{!}{\begin{tabular}{@{}l|l|c|c|c@{}}
\toprule
\textbf{Architecture}      & \textbf{Method} & \multicolumn{1}{l|}{\textbf{\begin{tabular}[c]{@{}l@{}}Aircraft\\ Flowers\\ STL10\end{tabular}}} & \multicolumn{1}{l|}{\textbf{\begin{tabular}[c]{@{}l@{}}GTSRB\\ Cars\\ Flowers\end{tabular}}} & \multicolumn{1}{l}{\textbf{\begin{tabular}[c]{@{}l@{}}STL10\\ Cars\\ GTSRB\end{tabular}}} \\ \midrule
\multirow{4}{*}{ResNet-18} & PAINT           & 20.02                                                                                            & 14.28                                                                                        & 26.85                                                                                     \\
                           & \methodname (1)        & 70.52                                                                                            & 64.95                                                                                        & 67.51                                                                                     \\
                           & \methodname (2)        & 71.93                                                                                            & 69.89                                                                                        & 72.37                                                                                     \\
                           & \methodname (4)        & \textbf{72.41}                                                                                   & \textbf{79.03}                                                                               & \textbf{81.60}                                                                            \\ \midrule
\multirow{4}{*}{ConvNeXT-Tiny} & PAINT           & 19.95                                                                                            & 28.68                                                                                        & 27.01                                                                                     \\
                           & \methodname (1)        & 83.52                                                                                            & 64.72                                                                                        & 65.74                                                                                     \\
                           & \methodname (2)        & \textbf{82.71}                                                                                   & \textbf{83.45}                                                                               & 85.94                                                                                     \\
                           & \methodname (4)        & 81.06                                                                                            & 83.04                                                                                        & \textbf{86.84}                                                                            \\ \bottomrule
\end{tabular}}
\vspace{-10pt}
\caption{Performance Evaluation of \methodname and PAINT\cite{modelpatching2022} in \textbf{\textit{Sequential Patching}} scenario. Mean of Top-1 accuracies is reported.}
\label{tab:seq-patching}
\vspace{-20pt}
\end{table}

\noindent \textbf{Sequential Patching:}
In this experimental configuration, tasks requiring patching are encountered sequentially. For PAINT, in accordance with \cite{modelpatching2022}, we commence with a pre-trained model, append a new fully connected (FC) layer, and proceed to fine-tune the backbone along with this new layer for the first task. The optimal interpolation coefficient is determined based on the harmonic mean of top-1 accuracies for both the pre-trained and newly patched tasks. These interpolated weights are then employed for subsequent tasks, and the best interpolation coefficient is recalibrated using the harmonic mean of accuracies across all preceding tasks. During evaluation, the task ID guides the selection of the appropriate FC layer.
\noindent For \methodname, a new \smethodname is introduced for each incoming task, in tandem with a new FC layer. These \smethodnames are trained alongside the FC layer for each individual task. Table \ref{tab:seq-patching} presents the average accuracy of the final patched model across all prior tasks. It's worth noting that we opt for average accuracy over harmonic mean in this setting due to the simultaneous evaluation of multiple tasks at the conclusion of the experiment.
\noindent Unlike PAINT, which does not preserve the base model and relies on interpolated models for future tasks, \methodname is designed to maintain consistent performance on previously learned tasks. A more comprehensive analysis of these findings is available in the Appendix.

\subsection{\methodname for Other Tasks and Settings}

\noindent We extend our investigation of \methodname to diverse scenarios, including transfer learning and zero-shot domain adaptation, as well as additional vision tasks like object detection and segmentation. Unlike previous patching efforts, which did not explore these settings, our aim is to demonstrate the versatility of \methodname in adapting pre-trained models across a range of tasks and conditions. While our focus is not on achieving state-of-the-art performance, we do illustrate how the incorporation of \smethodnames enhances the fine-tuning capabilities of existing pre-trained models.
\begin{table}[h]
\centering
\resizebox{\linewidth}{!}{\begin{tabular}{>{\kern-\tabcolsep}l|ccccr|ccccr<{\kern-\tabcolsep}}
\toprule
\textbf{Method}             & \multicolumn{5}{c|}{\textbf{FGVC Aircraft}}                                                       & \multicolumn{5}{c}{\textbf{Stanford Cars}}                                                       \\ \midrule
                   & 15\%          & 30\%          & 50\%          & 100\%         & \multicolumn{1}{l|}{Avg} & 15\%          & 30\%          & 50\%          & 100\%         & \multicolumn{1}{l}{Avg} \\ \midrule
\rowcolor{baseline} Vanilla Baseline          & 41.6          & 57.8          & 68.7          & 80.2          & 62.1                     & 41.1          & 65.9          & 78.4          & 87.8          & 68.3                    \\
+ LWF              & 44.1          & 60.6          & 68.7          & 82.4          & 64.0                     & 44.9          & 67.0          & 77.6          & 87.5          & 69.3                    \\
+ BSS              & 43.6          & 59.5          & 69.6          & 81.2          & 63.5                     & 43.3          & 67.6          & 79.6          & 88.0          & 69.6                    \\
+ DELTA            & 44.4          & 61.9          & 71.4          & 82.7          & 65.1                     & 45.0          & 68.4          & 79.6          & 88.4          & 70.4                    \\
+ Co-Tuning        & 45.9          & 61.2          & 71.3          & 82.2          & 65.2                     & 49.0          & 70.6          & 81.9          & 89.1          & 72.7                    \\
+ StochNorm        & 44.3          & 60.6          & 70.1          & 81.5          & 64.1                     & 44.4          & 68.1          & 79.1          & 87.9          & 69.9                    \\
+ Bi-Tuning        & \underline{47.2}          & 64.3          & 73.7          & 84.3          & 67.4                     & 48.3          & 72.8          & \underline{83.3}          & 90.2          & 73.7                    \\ \midrule
+ \methodname (4)             & 46.9          & \underline{64.8}          & \underline{74.3}          & \underline{85.3}          & \underline{67.8}                     & \underline{48.5}          & \underline{73.8}          & 82.5          & \underline{90.4}          & \underline{73.8}                    \\
\rowcolor{highlight} + Bi-Tuning + \methodname (4) & \textbf{47.7} & \textbf{65.1} & \textbf{76.5} & \textbf{87.1} & \textbf{69.1}            & \textbf{49.2} & \textbf{73.9} & \textbf{84.3} & \textbf{91.1} & \textbf{74.6}           \\ \bottomrule
\end{tabular}}
\vspace{-8pt}
\caption{\textbf{\textit{\methodname outperforms and enhances existing methods (see Bi-Tuning + \methodname)}}. Performance of \methodname vs. established methods on TL benchmark~\cite{dalib} under varied data availability (e.g., 50\% = half training data). \methodname (4) indicates use of four \smethodnames.}
\label{tab:compare-tl-approaches}
\vspace{-8pt}
\end{table}

\noindent \textbf{Transfer Learning (TL):} We study \methodname in a traditional TL setting and compare our methodology with six different transfer learning methods, including state-of-the-art approaches: LWF~\cite{LWF}, BSS~\cite{BSS}, DELTA~\cite{li2018delta}, StochNorm~\cite{stochnorm}, Co-Tuning~\cite{CoTuning}, and Bi-Tuning~\cite{bituning}. We also compare with a vanilla baseline, where the backbone architecture is fine-tuned end-to-end on the target dataset, and a combination of our method with Bi-Tuning~\cite{bituning}. We use the standard TL benchmark and training protocols provided in ~\cite{dalib}, comprising FGVC-Aircraft~\cite{fgvc-aircraft} and Stanford-Cars~\cite{stanford-cars} datasets, with ResNet-50 as the backbone, following~\cite{stochnorm,CoTuning,bituning}.To test the effectiveness of our approach in the low-data regime, we train the model at different levels, i.e. on 15\%, 30\%, 50\% and 100\% of the training data, for the above experiments. Table ~\ref{tab:compare-tl-approaches} presents our results.
\methodname outperforms existing approaches on almost all experiments consistently, even with varying amounts of training data in the target domain. Bi-Tuning ~\cite{bituning} uses contrastive learning and leverages both supervised and unsupervised pre-trained representations, which perhaps makes it stronger compared to other baselines. \methodname instead uses a simple strategy to outperform all baselines including Bi-Tuning on most settings. Moreover, adding  \methodname to Bi-Tuning helps achieve a new state-of-the-art on a standard TL benchmark~\cite{dalib}.

\begin{table}[h]
\centering
\resizebox{\linewidth}{!}{%
\begin{tabular}{>{\kern-\tabcolsep}l|ccc<{\kern-\tabcolsep}}
\toprule
\textbf{ResNet-18}                        & \textbf{MNIST $\rightarrow$USPS} & \textbf{USPS $\rightarrow$MNIST} & \textbf{SVHN$\rightarrow$MNIST} \\ \midrule
\rowcolor{baseline} Vanilla                       & 49.0                   & 42.81                   & 69.7                  \\
 + \methodname (4)                & 56.5                 & 47.0                   & 75.3                  \\
 + Uniform prior \cite{uniformprior}       & 67.2                  & 56.2                   & 71.3                  \\

\rowcolor{highlight} + Uniform prior + \methodname (4) & \textbf{71.3}           & \textbf{61.0 }           & \textbf{77.9}          \\ \midrule
\rowcolor{baseline} ADDA \cite{adda}                         & 88.2                  & 89.0                    & 73.4                  \\
  + \methodname(4)                   & 90.1                    & 90.7                   & 81.1                  \\
 + Uniform Prior          & 91.6                    & 92.7                   & 79.4                \\
\rowcolor{highlight} + Uniform Prior + \methodname(4)    & \textbf{92.4 }           & \textbf{94.1}           & \textbf{83.6 }          \\ \midrule
Target only                   & 98.1                    & 99.8                    & 99.8                 \\ \bottomrule
\end{tabular}
}
\vspace{-8pt}
\caption{\textit{\textbf{\methodname boosts ZSDA performance.}} \methodname (4) indicates four \smethodnames used.}
\vspace{-15pt}
\label{tab:zsda}
\end{table}

\noindent {\textbf{Zero-shot Domain Adaptation (ZSDA):}} 
In ZSDA, the models are trained on a given source dataset and evaluated on their ability to transfer to a different target dataset that has the same classes but is from a different data distribution.
We adopt the experimental framework outlined in \cite{uniformprior} to evaluate \methodname using a ResNet-18 architecture. Our comparison includes both the Uniform Prior method \cite{uniformprior} and Adversarial Discriminative Domain Adaptation (ADDA) \cite{adda}. Additionally, we explore the synergistic effects of combining our approach with these existing methods. The results are presented in Table \ref{tab:zsda}
We continue to observe that by adding \methodname, we can improve the network's performance on the target dataset significantly. Importantly, the results also show that \methodname can be a general strategy to use along with existing techniques.

\noindent \textbf{Object Detection and Segmentation: }
The utility of pre-trained ImageNet models extends beyond classification tasks; they serve as foundational backbones for object detection and semantic segmentation. To assess the efficacy of \methodname in these contexts, we integrate it into established architectures for object detection and semantic segmentation. Specifically, we employ the Faster-RCNN model \cite{renNIPS15fasterrcnn} with ResNet-18 and ResNet-34 backbones for object detection, and DeepLab-V3 \cite{deeplabv3} with a ResNet-50 backbone for semantic segmentation. All experiments are conducted on the PASCAL-VOC dataset \cite{pascal-voc-2007}.

\noindent In these experiments, we introduce \methodname into the backbone networks and train only the \smethodnames, keeping the backbone parameters frozen. The results, summarized in Table \ref{tab:detect-segm-results}, demonstrate that the inclusion of \methodname leads to performance gains in both object detection and semantic segmentation tasks.

\begin{table}
\centering
\resizebox{0.7\linewidth}{!}{\begin{tabular}{>{\kern-\tabcolsep}c|l|c<{\kern-\tabcolsep}}
\toprule
Task                                                                              & Backbone        & mAP/IoU        \\ \midrule
\multirow{4}{*}{\begin{tabular}[c]{@{}c@{}}Object \\ Detection\end{tabular}}                                                 & ResNet-18       & 59.72         \\
                                                                                  & \cellcolor{highlight} + \methodname (4) &\cellcolor{highlight} \textbf{61.33} \\ \cmidrule(l){2-3} 
                                                                                  & ResNet-34       & 64.40         \\
                                                                                  & \cellcolor{highlight} + \methodname (4) &\cellcolor{highlight} \textbf{67.80} \\ \midrule
\multirow{2}{*}{\begin{tabular}[c]{@{}c@{}}Semantic\\ Segementation\end{tabular}} & ResNet-50       & 70.02          \\
                                                                                  & \cellcolor{highlight}+ \methodname (4) &\cellcolor{highlight} \textbf{70.80 } \\ \bottomrule
\end{tabular}}
\vspace{-7pt}
\caption{\textit{\textbf{\methodname enhances performance on complex tasks like detection and segmentation.}} Results on Faster-RCNN and DeepLab-v3 with various backbones. \methodname (4) indicates four \smethodnames used.}
\vspace{-20pt}
\label{tab:detect-segm-results}
\end{table}

\subsection{What makes \methodname work? Analysis and Ablation Studies}
\noindent  We herein study the importance of the components of \methodname -- in particular, the use of skip connections, both learnable and input-conditioned, as well as the relevance of batch normalization layers. 
We consider four variants of ResNets, including additional architectures beyond what we studied so far (for analysis purposes), and train them in multiple settings on two datasets namely, FGVC-Aircraft \cite{fgvc-aircraft} and Stanford-Cars \cite{stanford-cars}.

\begin{table}[]
\resizebox{\linewidth}{!}{\begin{tabular}{>{\kern-\tabcolsep}l|cc<{\kern-\tabcolsep}}
\toprule
\textbf{Architecture}                       & \textbf{FGVC Aircraft}  & \textbf{Stanford Cars}  \\ \midrule
\rowcolor{baseline}ResNet-18 + FC (R-18)      & 59.89          & 67.63          \\
+ Learnable Skip connections       & 59.26          & 51.71          \\ \cmidrule(r){1-1}
R-18 + All Skips (Learnable)       & 63.70           & 59.85          \\
+ BN                               & 70.69          & \textbf{77.63} \\ \cmidrule(r){1-1}
R-18 + All Skips                   & 55.18          & 61.43          \\
+ BN                               & \underline{71.12}          & 75.43          \\
+ Random                           & 21.33          & 26.38          \\
\rowcolor{highlight}+ \methodname (4)                       & \textbf{72.91} & \underline{77.21}          \\ \midrule
\rowcolor{baseline}ResNet-34 + FC (R-34)       & 59.83          & 68.9           \\
+ Learnable Skip connections       & 61.48          & 65.26          \\ \cmidrule(r){1-1}
R-34 + All Skips (Learnable)       & 41.52          & 27.91          \\
+ BN                               & 64.63          & \underline{79.31}          \\ \cmidrule(r){1-1}
R34 + All Skips                    & 47.34          & 44.45          \\
+ BN                               & \underline{67.78}          & 66.19          \\
+ Random                           & 9.31           & 7.16           \\
\rowcolor{highlight}+ \methodname (4)                      & \textbf{75.19} & \textbf{81.51} \\ \midrule
\rowcolor{baseline}ResNeXt-50 + FC (RX-50)     & 61.06          & 69.39          \\
 + Learnable Skip connections        & 58.87          & 69.23          \\ \cmidrule(r){1-1}
RX-50 + All Skips (Learnable)        & 47.53          & 72.49          \\
+ BN                               & 50.85          & \underline{77.04}          \\ \cmidrule(r){1-1}
RX-50 + All Skips                   & 52.72          & 50.14          \\
+ BN                               & \underline{72.82}          & 75.56          \\
+ Random                           & 20.94          & 21.81          \\
\rowcolor{highlight}+\methodname (4)                       & \textbf{78.04} & \textbf{85.25} \\ \midrule
\rowcolor{baseline} WideResNet-50 + FC (WRN-50)  & 55.96          & 64.89          \\
+ Learnable Skip connections     & 56.55          & 64.45          \\ \cmidrule(r){1-1}
WRN-50 + All Skips (Learnable)     & 30.92          & 60.47          \\
+ BN                               & 53.51          & \underline{76.40}           \\ \cmidrule(r){1-1}
WRN-50 + All Skips               & 51.28          & 48.97          \\
+ BN                               & \underline{72.52}          & 75.15          \\
+ Random                           & 14.37          & 12.05          \\
\rowcolor{highlight}+ \methodname (4)                       & \textbf{76.18} & \textbf{83.08} \\ \bottomrule
\end{tabular}}
\vspace{-7pt}
\caption{Quantitative ablation analysis of marginal contribution of each component in \methodname. \methodname (4) denotes the use of four \smethodnames}
\vspace{-20pt}
\label{tab:ablation}
\end{table}

\begin{table*}[h]
\centering
\footnotesize
\setlength{\extrarowheight}{0pt}
\addtolength{\extrarowheight}{\aboverulesep}
\setlength{\aboverulesep}{0pt}
\setlength{\belowrulesep}{0pt}
\label{tab:comparison}
\begin{tabular}{l|c|c|c|c} 
\toprule
\rowcolor[rgb]{0.902,0.898,0.902} \begin{tabular}[c]{@{}>{\cellcolor[rgb]{0.902,0.898,0.902}}l@{}}\textsc{~~Characteristics~$\rightarrow$}\\\textit{}\\\textsc{Settings~$\downarrow$}\end{tabular} & \begin{tabular}[c]{@{}>{\cellcolor[rgb]{0.902,0.898,0.902}}c@{}}\textit{\textbf{Adapt off-the-shelf }}\\\textit{\textbf{pre-trained models}}\end{tabular} & \begin{tabular}[c]{@{}>{\cellcolor[rgb]{0.902,0.898,0.902}}c@{}}\textit{\textbf{Maintain performance }}\\\textit{\textbf{on base task}}\end{tabular} & \begin{tabular}[c]{@{}>{\cellcolor[rgb]{0.902,0.898,0.902}}c@{}}\textbf{\textit{Adapt to multiple }}\\\textbf{\textit{tasks overtime}}\end{tabular} & \begin{tabular}[c]{@{}>{\cellcolor[rgb]{0.902,0.898,0.902}}c@{}}\textit{\textbf{End-to-end}}\\\textit{\textbf{learning}}\end{tabular}  \\ 
\hline
\textcolor{uniform}{Meta Learning~\cite{hospedales2021metalearning-survey}}
    & \textcolor{uniform}{\xmark}
    & \textcolor{uniform}{\xmark}
    & \textcolor{uniform}{\cmarkcircle}
    & \textcolor{uniform}{\cmark}                                                                                                                  \\
 \textcolor{uniform}{Zero/Few-shot Learning~\cite{song2022comprehensive-fslsurvery, bendre2020learning-zsl-survey}}                                                                                                           & \textcolor{uniform}{\xmark}                                                                                                                        & \textcolor{uniform}{\cmark}                                                                                                                  & \textcolor{uniform}{\xmark}                                                                                                                  & \textcolor{uniform}{\cmark}                                                                                                     \\
\textcolor{uniform}{Continual Learning~\cite{shaheen2022continual-survey, qu2021recent-clsurvey}}                                                                                                               & \textcolor{uniform}{\xmark}                                                                                                                        & \textcolor{uniform}{\cmarkcircle}                                                                                                                  & \textcolor{uniform}{\cmark}                                                                                                                 & \textcolor{uniform}{\cmark}                                                                                                     \\ 
\hline
 Transfer Learning~\cite{tl-survey}                                                                                                                                                & \cmark                                                                                                                                                       & \xmark                                                                                                                                                   & \cmarkcircle                                                                                                                                                 & \cmark                                                                                                                                     \\
 Model Editing/Debugging~\cite{sinitsin2020editable, ribeiro2022adaptive}                                                                                                                                          & \cmark                                                                                                                                                       & \cmark                                                                                                                                                  & \xmark                                                                                                                                                  & \cmark                                                                                                                                     \\
Patching~\cite{modelpatching2022}                                                                                                                                                         & \cmark                                                                                                                                                       & \cmarkcircle                                                                                                                                                  & \cmark                                                                                                                                                 & \xmark                                                                                                                                      \\
\rowcolor{highlight} \settingname (proposed method)                                                                                                                                                  & \cmark                                                                                                                                                       & \cmark                                                                                                                                                  & \cmark                                                                                                                                                 & \cmark                                                                                                                                     \\
\bottomrule
\end{tabular}
\vspace{-5pt}
\caption{Comparison of proposed \settingname with other related settings. \cmark: Yes, \xmark: No, \cmarkcircle: Yes in certain cases.
Similar to \cite{modelpatching2022}, we focus on adapting off-the-shelf pre-trained models in this work, unlike zero/few-shot learning or continual learning.}
\label{tab:comparison}
\vspace{-17pt}
\end{table*}

\noindent \textbf{FC fine-tuning as a baseline: }
A compelling baseline for the patching process is the fine-tuning of just the fully-connected (FC) layer. As elaborated in Section \ref{sec_method}, the addition of a new FC layer is an unavoidable step when patching CNNs. Given that FC-based fine-tuning does not necessitate any modifications to the backbone architecture, it serves as an ideal starting point for our ablation studies. This approach allows us to isolate and understand the impact of each component introduced in \methodname. By comparing the performance of \methodname against this baseline, we can more precisely attribute any performance gains to the specific elements of our method.

\noindent \textbf{Role of Skip Connections and Input-Conditioning.} In order to study this, we introduce four kinds of skip connections in the standard ResNets. (1) \textit{Learnable Skip Connections:} We make the already existing skip connections in residual networks soft by introducing a weight parameter, i.e. when adding an input to the output block, instead of using a simple identity mapping, we use a scaled identity mapping (multiply the input using this learnable weight parameter). (2) \textit{All Skips:} We introduce all possible skip connections inside each ResNet module. (3) \textit{All Skips (Learnable):}  All Skip connections with learnable weights i.e., after all possible skip connections are introduced, we  assign a separate weight to all of them and learn the weight using standard backpropagation. (4) \textit{Skip Connections in \methodname:} When a \smethodname is used, we introduce all possible skip connections and weight those skip connections in an input-conditioned manner i.e., the weight of the skip connection is determined by the \smethodname. In (1) and (3), the skip connections weights are not input-conditioned; once learned on the training set, they stay fixed during inference irrespective of the input. 
Table \ref{tab:ablation} shows the results of this study on patching. Evidently, \methodname outperform other strategies in the results. %
Introducing all possible skip connections without input-conditioning does not help much. While learnable skip connections seems to help marginally, the improvement is low compared to \methodname-based skip connections which are input-conditioned.

\noindent \textbf{Role of Batch Normalization (BN).} We explicitly study the role of BN layers since introducing all possible skip connections can result in an increased variance. We introduce a BN layer in our approach to curtail this.
Table \ref{tab:ablation} presents related results; as seen in the table, adding BN layers improves performance in general even for other approaches when skip connections are added, demonstrating their usefulness in  approaches that modify or adapt skip connections. 

\vspace{-10pt}

\section{Related Work}
\label{sec:related-work}
\vspace{-5pt}
\noindent We categorize the related work into two overarching themes: settings and methods. This organization allows us to delineate the unique aspects of our proposed setting, \settingname, and method, \methodname, in relation to existing paradigms.

\noindent \textbf{Related Settings}
We provide a comparative analysis of various problem settings, as summarized in Table \ref{tab:comparison} and Fig \ref{fig:teaser}, to contextualize \settingname

\noindent {\textit{Meta-Learning (ML):}} ML focuses on training a meta-model that generalizes across tasks \cite{finn2017maml,hospedales2021metalearning-survey}. Unlike 
\settingname, meta-learning does not aim to preserve primary task performance while adapting to new tasks.

\noindent \textit{{Zero-shot/Few-shot learning (ZSL/FSL):}}  These paradigms aim for generalization with limited labeled data for new tasks \cite{song2022comprehensive-fslsurvery, bendre2020learning-zsl-survey}. \settingname diverges by enabling multi-task fine-tuning without altering the base model.

\noindent {\textit{Continual Learning (CL):}}  CL methods incrementally adapt to new tasks while mitigating catastrophic forgetting \cite{shaheen2022continual-survey,qu2021recent-clsurvey}. In contrast, \settingname employs a static, large-scale pre-trained model for rapid task adaptation without model modification. The primary objective of \settingname is quick adaptation of a pre-trained model for a given task \textit{{without modifying the original model}}, while IL methods \textit{{incrementally modify a model}} being performant on all tasks introduced to it. 

\noindent {\textit{Transfer Learning (TL):}} TL methods fine-tune pre-trained models for new tasks \cite{tl-survey,plested2022deep}. Unlike TL, \settingname explicitly maintains base task performance, allowing for multi-task support with a single model instance. 

\noindent {\textit{Model Editing/Debugging:}} These methods modify models at the sample level to correct or update predictions \cite{sinitsin2020editable,de2021editing,mitchell2021fast,shibani2021editing,ribeiro2020beyond,ribeiro2022adaptive}. \settingname, however, operates at the task or dataset level.

\noindent \textbf{Related Methods.} We now shift our focus to methods that share similarities or objectives with our proposed approach \methodname.

\noindent {\textit{Model Patching:}} Recent works have explored patching pre-trained models for specific objectives \cite{goel2021model,modelpatching2022}. While our work aligns more with the goals of \cite{modelpatching2022}, our approach and formulation, as detailed in Sections \ref{sec_intro} and \ref{sec_method}, are very distinct.

\noindent {\textit{Use of Skip Connections:}} Skip connections have been a subject of extensive research since their introduction \cite{resnets2015}. Variants like DenseNets \cite{densenets2016} and SparseNets \cite{sparsenet} have explored different configurations of skip connections. Other works like \cite{wscisr} have introduced weighted skip connections for specific applications like super-resolution. However, none of these works leverage skip connections for the purpose of multi-task learning or patching, which is the primary focus of this work.

\noindent {\textit{Input-conditioned architectures:}} Several methods have been proposed to adapt neural network architectures based on the input \cite{SkipNet,batch-shaping,blockdrop,networks-with-stochastic-depth,pathnet} with the primary goal of efficient inference. SpotTune~\cite{spottune}, for instance, fine-tunes layers in an input-conditioned manner primarily for transfer learning. While \methodname also adapts to each input, it does so uniquely by focusing on input-adaptive skip connections, setting it apart from methods that adapt at the layer or block level. To ensure a fair comparison, we compare our method with the most recent transfer learning works in Table ~\ref{tab:compare-tl-approaches}.

\vspace{-6pt}
\section{Conclusions and Future Work}
\vspace{-4pt}
\noindent In this work, we propose a new setting \settingname, to extend the definition of model patching \cite{modelpatching2022} to general purpose CNNs, and also ensuring the maintenance of performance on a base task. We propose \methodname as a simple and lightweight architectural modification to efficiently implement our model patching variant in this work. \settingname significantly outperforms existing patching methods while only using a fraction of the parameters for training. We also show that the proposed approach can be used for other problem settings such as transfer learning and zero-shot domain adaptation, as well as other vision tasks such as object detection and segmentation. Future directions include extensions to large-scale vision-language models and simultaneous adaptation to multiple tasks while only training a fraction of parameters. 

\vspace{-10pt}
{\small
\bibliographystyle{ieee_fullname}
\bibliography{egbib}

\begin{thebibliography}{10}\itemsep=-1pt

\bibitem{hugging-face-2022}
{Pre-trained models for Image Classification}, 01 2023.
\newblock \url{https://huggingface.co/models?pipeline_tag=image-classification&sort=downloads}.

\bibitem{batch-shaping}
Babak~Ehteshami Bejnordi, Tijmen Blankevoort, and Max Welling.
\newblock Batch-shaping for learning conditional channel gated networks.
\newblock In {\em International Conference on Learning Representations}, 2020.

\bibitem{bendre2020learning-zsl-survey}
Nihar Bendre, Hugo~Terashima Marín, and Peyman Najafirad.
\newblock Learning from few samples: A survey.
\newblock {\em arXiv preprint arXiv: Arxiv-2007.15484}, 2020.

\bibitem{deeplabv3}
Liang-Chieh Chen, Yukun Zhu, George Papandreou, Florian Schroff, and Hartwig Adam.
\newblock Encoder-decoder with atrous separable convolution for semantic image segmentation.
\newblock In {\em Proceedings of the European Conference on Computer Vision (ECCV)}, September 2018.

\bibitem{BSS}
Xinyang Chen, Sinan Wang, Bo Fu, Mingsheng Long, and Jianmin Wang.
\newblock Catastrophic forgetting meets negative transfer: Batch spectral shrinkage for safe transfer learning.
\newblock In H. Wallach, H. Larochelle, A. Beygelzimer, F. d\textquotesingle Alch\'{e}-Buc, E. Fox, and R. Garnett, editors, {\em Advances in Neural Information Processing Systems}, volume~32. Curran Associates, Inc., 2019.

\bibitem{de2021editing}
Nicola De~Cao, Wilker Aziz, and Ivan Titov.
\newblock Editing factual knowledge in language models.
\newblock In {\em Conference on Empirical Methods in Natural Language Processing (EMNLP)}, 2021.

\bibitem{pascal-voc-2007}
M. Everingham, L. Van~Gool, C.~K.~I. Williams, J. Winn, and A. Zisserman.
\newblock The {PASCAL} {V}isual {O}bject {C}lasses {C}hallenge 2007 {(VOC2007)} {R}esults.
\newblock http://www.pascal-network.org/challenges/VOC/voc2007/workshop/index.html.

\bibitem{pathnet}
Chrisantha {Fernando}, Dylan {Banarse}, Charles {Blundell}, Yori {Zwols}, David {Ha}, Andrei~A. {Rusu}, Alexander {Pritzel}, and Daan {Wierstra}.
\newblock {PathNet: Evolution Channels Gradient Descent in Super Neural Networks}.
\newblock {\em arXiv e-prints}, page arXiv:1701.08734, Jan. 2017.

\bibitem{finn2017maml}
Chelsea Finn, Pieter Abbeel, and Sergey Levine.
\newblock Model-agnostic meta-learning for fast adaptation of deep networks.
\newblock In {\em Proceedings of the 34th International Conference on Machine Learning}, pages 1126--1135, 2017.

\bibitem{GlorotAISTATS2010}
Xavier Glorot and Yoshua Bengio.
\newblock Understanding the difficulty of training deep feedforward neural networks.
\newblock In {\em JMLR W\&CP: Proceedings of the Thirteenth International Conference on Artificial Intelligence and Statistics (AISTATS 2010)}, volume~9, pages 249--256, May 2010.

\bibitem{goel2021model}
Karan Goel, Albert Gu, Yixuan Li, and Christopher Re.
\newblock Model patching: Closing the subgroup performance gap with data augmentation.
\newblock In {\em International Conference on Learning Representations}, 2021.

\bibitem{spottune}
Yunhui Guo, Honghui Shi, Abhishek Kumar, Kristen Grauman, Tajana Rosing, and Rogerio Feris.
\newblock Spottune: Transfer learning through adaptive fine-tuning.
\newblock In {\em Proceedings of the IEEE/CVF Conference on Computer Vision and Pattern Recognition (CVPR)}, June 2019.

\bibitem{hypernetwork2016}
David Ha, Andrew~M. Dai, and Quoc~V. Le.
\newblock Hypernetworks.
\newblock In {\em International Conference on Learning Representations}, 2017.

\bibitem{heinit}
Kaiming He, Xiangyu Zhang, Shaoqing Ren, and Jian Sun.
\newblock Delving deep into rectifiers: Surpassing human-level performance on imagenet classification.
\newblock In {\em 2015 IEEE International Conference on Computer Vision (ICCV)}, pages 1026--1034, 2015.

\bibitem{resnets2015}
Kaiming He, Xiangyu Zhang, Shaoqing Ren, and Jian Sun.
\newblock Deep residual learning for image recognition.
\newblock In {\em Proceedings of the IEEE Conference on Computer Vision and Pattern Recognition (CVPR)}, June 2016.

\bibitem{hospedales2021metalearning-survey}
Timothy Hospedales, Antreas Antoniou, Paul Micaelli, and Amos Storkey.
\newblock Meta-learning in neural networks: A survey.
\newblock {\em IEEE transactions on pattern analysis and machine intelligence}, 44(9):5149--5169, 2021.

\bibitem{densenets2016}
Gao Huang, Zhuang Liu, Laurens van~der Maaten, and Kilian~Q. Weinberger.
\newblock Densely connected convolutional networks.
\newblock In {\em Proceedings of the IEEE Conference on Computer Vision and Pattern Recognition (CVPR)}, July 2017.

\bibitem{networks-with-stochastic-depth}
Gao Huang, Yu Sun, Zhuang Liu, Daniel Sedra, and Kilian~Q. Weinberger.
\newblock Deep networks with stochastic depth.
\newblock In Bastian Leibe, Jiri Matas, Nicu Sebe, and Max Welling, editors, {\em Computer Vision -- ECCV 2016}, pages 646--661, Cham, 2016. Springer International Publishing.

\bibitem{modelpatching2022}
Gabriel Ilharco, Mitchell Wortsman, Samir~Yitzhak Gadre, Shuran Song, Hannaneh Hajishirzi, Simon Kornblith, Ali Farhadi, and Ludwig Schmidt.
\newblock Patching open-vocabulary models by interpolating weights.
\newblock In {\em Advances in Neural Information Processing Systems}, 2022.

\bibitem{dalib}
Jiang {Junguang}, {Baixu} Chen, Fu {Bo}, and {Mingsheng} Long.
\newblock Transfer-learning-library.
\newblock \url{https://github.com/thuml/Transfer-Learning-Library}, 2020.

\bibitem{stochnorm}
Zhi Kou, Kaichao You, Mingsheng Long, and Jianmin Wang.
\newblock Stochastic normalization.
\newblock In H. Larochelle, M. Ranzato, R. Hadsell, M.~F. Balcan, and H. Lin, editors, {\em Advances in Neural Information Processing Systems}, volume~33, pages 16304--16314. Curran Associates, Inc., 2020.

\bibitem{stanford-cars}
Jonathan Krause, Michael Stark, Jia Deng, and Li Fei-Fei.
\newblock 3d object representations for fine-grained categorization.
\newblock In {\em 4th International IEEE Workshop on 3D Representation and Recognition (3dRR-13)}, Sydney, Australia, 2013.

\bibitem{li2018delta}
Xingjian Li, Haoyi Xiong, Hanchao Wang, Yuxuan Rao, Liping Liu, and Jun Huan.
\newblock {DELTA}: {DEEP} {LEARNING} {TRANSFER} {USING} {FEATURE} {MAP} {WITH} {ATTENTION} {FOR} {CONVOLUTIONAL} {NETWORKS}.
\newblock In {\em International Conference on Learning Representations}, 2019.

\bibitem{LWF}
Zhizhong Li and Derek Hoiem.
\newblock Learning without forgetting.
\newblock In {\em ECCV}, 2016.

\bibitem{sparsenet}
Wenqi {Liu} and Kun {Zeng}.
\newblock {SparseNet: A Sparse DenseNet for Image Classification}.
\newblock {\em arXiv e-prints}, page arXiv:1804.05340, Apr. 2018.

\bibitem{fgvc-aircraft}
S. Maji, J. Kannala, E. Rahtu, M. Blaschko, and A. Vedaldi.
\newblock Fine-grained visual classification of aircraft.
\newblock Technical report, 2013.

\bibitem{mirza2022norm}
M~Jehanzeb Mirza, Jakub Micorek, Horst Possegger, and Horst Bischof.
\newblock The norm must go on: dynamic unsupervised domain adaptation by normalization.
\newblock In {\em CVPR}, 2022.

\bibitem{mitchell2021fast}
Eric Mitchell, Charles Lin, Antoine Bosselut, Chelsea Finn, and Christopher~D Manning.
\newblock Fast model editing at scale.
\newblock In {\em International Conference on Learning Representations (ICLR)}, 2021.

\bibitem{flowers-102}
M-E. Nilsback and A. Zisserman.
\newblock Automated flower classification over a large number of classes.
\newblock In {\em Proceedings of the Indian Conference on Computer Vision, Graphics and Image Processing}, Dec 2008.

\bibitem{paszke2019pytorch}
Adam Paszke, Sam Gross, Francisco Massa, Adam Lerer, James Bradbury, Gregory Chanan, Trevor Killeen, Zeming Lin, Natalia Gimelshein, Luca Antiga, Alban Desmaison, Andreas Köpf, Edward Yang, Zach DeVito, Martin Raison, Alykhan Tejani, Sasank Chilamkurthy, Benoit Steiner, Lu Fang, Junjie Bai, and Soumith Chintala.
\newblock Pytorch: An imperative style, high-performance deep learning library.
\newblock {\em arXiv preprint arXiv: Arxiv-1912.01703}, 2019.

\bibitem{plested2022deep}
Jo Plested and Tom Gedeon.
\newblock Deep transfer learning for image classification: a survey.
\newblock {\em arXiv preprint arXiv: Arxiv-2205.09904}, 2022.

\bibitem{qu2021recent-clsurvey}
Haoxuan Qu, Hossein Rahmani, Li Xu, Bryan Williams, and Jun Liu.
\newblock Recent advances of continual learning in computer vision: An overview.
\newblock {\em arXiv preprint arXiv: Arxiv-2109.11369}, 2021.

\bibitem{pmlr-v139-radford21a}
Alec Radford, Jong~Wook Kim, Chris Hallacy, Aditya Ramesh, Gabriel Goh, Sandhini Agarwal, Girish Sastry, Amanda Askell, Pamela Mishkin, Jack Clark, Gretchen Krueger, and Ilya Sutskever.
\newblock Learning transferable visual models from natural language supervision.
\newblock In Marina Meila and Tong Zhang, editors, {\em Proceedings of the 38th International Conference on Machine Learning}, volume 139 of {\em Proceedings of Machine Learning Research}, pages 8748--8763. PMLR, 18--24 Jul 2021.

\bibitem{ml-like-software-2021}
Colin Raffel.
\newblock {A call to build models like we build opensource software}, 2021.
\newblock \url{https://cutt.ly/d8Jbq4J}.

\bibitem{Rebuffi17}
S-A Rebuffi, H. Bilen, and A. Vedaldi.
\newblock Learning multiple visual domains with residual adapters.
\newblock In {\em NeurIPS}, 2017.

\bibitem{rebuffi-cvpr2018}
Sylvestre-Alvise Rebuffi, Hakan Bilen, and Andrea Vedaldi.
\newblock Efficient parametrization of multi-domain deep neural networks.
\newblock In {\em CVPR}, 2018.

\bibitem{renNIPS15fasterrcnn}
Shaoqing Ren, Kaiming He, Ross Girshick, and Jian Sun.
\newblock Faster {R-CNN}: Towards real-time object detection with region proposal networks.
\newblock In {\em Advances in Neural Information Processing Systems ({NIPS})}, 2015.

\bibitem{ribeiro2022adaptive}
Marco~Tulio Ribeiro and Scott Lundberg.
\newblock Adaptive testing and debugging of nlp models.
\newblock In {\em Association for Computational Linguistics (ACL)}, 2022.

\bibitem{ribeiro2020beyond}
Marco~Tulio Ribeiro, Tongshuang Wu, Carlos Guestrin, and Sameer Singh.
\newblock Beyond accuracy: Behavioral testing of {NLP} models with {C}heck{L}ist.
\newblock In {\em Association for Computational Linguistics (ACL)}, 2020.

\bibitem{shibani2021editing}
Shibani Santurkar, Dimitris Tsipras, Mahalaxmi Elango, David Bau, Antonio Torralba, and Aleksander Madry.
\newblock Editing a classifier by rewriting its prediction rules.
\newblock In {\em Advances in Neural Information Processing Systems (NeurIPS)}, 2021.

\bibitem{shaheen2022continual-survey}
Khadija Shaheen, Muhammad~Abdullah Hanif, Osman Hasan, and Muhammad Shafique.
\newblock Continual learning for real-world autonomous systems: Algorithms, challenges and frameworks.
\newblock {\em Journal of Intelligent \& Robotic Systems}, 105(1):1--32, 2022.

\bibitem{NEURIPS2020_9b861925}
Jie Shao, Kai Hu, Changhu Wang, Xiangyang Xue, and Bhiksha Raj.
\newblock Is normalization indispensable for training deep neural network?
\newblock In H. Larochelle, M. Ranzato, R. Hadsell, M.~F. Balcan, and H. Lin, editors, {\em Advances in Neural Information Processing Systems}, volume~33, pages 13434--13444. Curran Associates, Inc., 2020.

\bibitem{uniformprior}
Samarth Sinha, Karsten Roth, Anirudh Goyal, Marzyeh Ghassemi, Zeynep Akata, Hugo Larochelle, and Animesh Garg.
\newblock Uniform priors for data-efficient learning.
\newblock In {\em Proceedings of the IEEE/CVF Conference on Computer Vision and Pattern Recognition (CVPR) Workshops}, pages 4017--4028, June 2022.

\bibitem{sinitsin2020editable}
Anton Sinitsin, Vsevolod Plokhotnyuk, Dmitriy Pyrkin, Sergei Popov, and Artem Babenko.
\newblock Editable neural networks.
\newblock In {\em International Conference on Learning Representations (ICLR)}, 2020.

\bibitem{song2022comprehensive-fslsurvery}
Yisheng Song, Ting Wang, Subrota~K Mondal, and Jyoti~Prakash Sahoo.
\newblock A comprehensive survey of few-shot learning: Evolution, applications, challenges, and opportunities.
\newblock {\em arXiv preprint arXiv: Arxiv-2205.06743}, 2022.

\bibitem{adda}
Eric Tzeng, Judy Hoffman, Kate Saenko, and Trevor Darrell.
\newblock Adversarial discriminative domain adaptation.
\newblock In {\em Proceedings of the IEEE Conference on Computer Vision and Pattern Recognition (CVPR)}, July 2017.

\bibitem{wscisr}
Jiachen Wang and Yingyun Yang.
\newblock Sellf-adaptive weighted skip connections for image super-resolution.
\newblock In {\em 2020 International Conference on Culture-oriented Science Technology (ICCST)}, pages 192--197, 2020.

\bibitem{SkipNet}
Xin Wang, Fisher Yu, Zi-Yi Dou, Trevor Darrell, and Joseph~E. Gonzalez.
\newblock Skipnet: Learning dynamic routing in convolutional networks.
\newblock In {\em Proceedings of the European Conference on Computer Vision (ECCV)}, September 2018.

\bibitem{rw2019timm}
Ross Wightman.
\newblock Pytorch image models, 2019.

\bibitem{blockdrop}
Zuxuan Wu, Tushar Nagarajan, Abhishek Kumar, Steven Rennie, Larry~S Davis, Kristen Grauman, and Rogerio Feris.
\newblock Blockdrop: Dynamic inference paths in residual networks.
\newblock In {\em CVPR}, 2018.

\bibitem{CoTuning}
Kaichao You, Zhi Kou, Mingsheng Long, and Jianmin Wang.
\newblock Co-tuning for transfer learning.
\newblock In H. Larochelle, M. Ranzato, R. Hadsell, M.~F. Balcan, and H. Lin, editors, {\em Advances in Neural Information Processing Systems}, volume~33, pages 17236--17246. Curran Associates, Inc., 2020.

\bibitem{fixupinit}
Hongyi Zhang, Yann~N. Dauphin, and Tengyu Ma.
\newblock Residual learning without normalization via better initialization.
\newblock In {\em International Conference on Learning Representations}, 2019.

\bibitem{bituning}
Jincheng {Zhong}, Ximei {Wang}, Zhi {Kou}, Jianmin {Wang}, and Mingsheng {Long}.
\newblock {Bi-tuning of Pre-trained Representations}.
\newblock {\em arXiv e-prints}, page arXiv:2011.06182, Nov. 2020.

\bibitem{tl-survey}
Fuzhen Zhuang, Zhiyuan Qi, Keyu Duan, Dongbo Xi, Yongchun Zhu, Hengshu Zhu, Hui Xiong, and Qing He.
\newblock A comprehensive survey on transfer learning.
\newblock {\em Proceedings of the IEEE}, PP:1--34, 07 2020.

\end{thebibliography}
}

\section*{\centering Appendix}

\noindent In this appendix, we provide additional details which we could not include in the main paper due to space constraints, including some quantitative and qualitative
results, as well as corresponding discussion and analysis that provide more insights into the proposed method.

\section{ Theoretical Analysis of Variance Issues} \label{sec:theory}
\vspace{-3pt}

\noindent As mentioned towards the end of Section ~\ref{sec_method},
with \methodname, when all possible skip
connections are introduced, the variance increases rapidly
since we add features from all preceding blocks. We address this issue using one BN layer in each \smethodname in our implementation. In this section, we discuss the reasons for increase in variance in detail, and prove it formally.

\noindent Weight initialization methods like He's initialization \cite{heinit} are quite commonly used in all CNN architectures to maintain the variance between input and output of every block or layer. 
However, when skip connections are present in an architecture, such initialization methods \cite{heinit} result in increased variance as shown in \cite{fixupinit}.
To understand this problem better, we look at  the output of a given $j$th block, $\mathcal{F}^j$, with one skip connection.
The output representation is then given by: $\mathcal{F}^j = \mathcal{F}^j(\mathcal{F}^{j-1}(\mathbf{x})) + \mathcal{F}^{j-1}(\mathbf{x})$. The variance of the output representation can hence be written as: 
\vspace{-6pt}
\begin{equation}
\begin{aligned}
    \operatorname{Var}\Big[\mathcal{F}^j\Big] & = \operatorname{Var}\Big[\mathcal{F}^j(\mathcal{F}^{j-1}(\mathbf{x}) + \mathcal{F}^{j-1}(\mathbf{x})\Big] \\ 
    &= \operatorname{Var}\Big[\mathcal{F}^j(\mathcal{F}^{j-1}(\mathbf{x}))\Big] +  \operatorname{Var}\Big[\mathcal{F}^{j-1}(\mathbf{x}))\Big] \\ & \quad \quad + \operatorname{Cov}\Big[\mathcal{F}^j(\mathcal{F}^{j-1}(\mathbf{x})), \mathcal{F}^{j-1}(\mathbf{x}))\Big] \\
    \label{eq:6}
\end{aligned}
\end{equation}

\vspace{-0.7cm}
\noindent where the first two terms refer to variance of input and output blocks respectively, and the third term denotes the covariance between  input and output. While initialization methods maintain the variance between input and output of a block \cite{heinit},
when skip connections are present, the overall variance of the block output is approximately two times the variance of the input.
\vspace{-6pt}
\begin{equation}\label{eq:7}
    \operatorname{Var}\Big[\mathcal{F}^j\Big] \approx 2\operatorname{Var} \Big[\mathcal{F}^{j-1}(\mathbf{x})\Big] 
    \vspace{-5pt}
\end{equation}
\noindent To show this formally, we follow two observations: (1) When Kaiming initialization is used, the output variance of a residual block (without skip connection) is (approximately) equal to the input variance i.e., $\operatorname{Var}(\mathcal{F}^{j-1}) (\mathbf{x}) \approx \operatorname{Var}(\mathbf{x})$; and (2) The input and output of a residual block are weakly correlated (as also suggested in \cite{NEURIPS2020_9b861925}).

\noindent \textbf{(1) Output Variance $\approx$ Input Variance:} With Kaiming initialization, $\operatorname{Var}(\mathcal{F}^{j-1}(\mathbf{x})) \approx \operatorname{Var}(\mathbf{x})$ \cite{heinit}. Consider the variance at a particular convolutional layer of a neural network, whose output is given by: %
\begin{equation}\label{eq:forward}
  \ve{y}_l=\ma{W}_l\ve{x}_l + \ve{b}_l
\end{equation}
\noindent where $\ve{x}$ is a $k^2c \times 1$ vector that represents $k \times k$ pixels in $c$ input channels, and $k$ is the spatial filter size. With $n=k^2c$ denoting the number of connections of a response, 
$\ma{W}$ is a $d$-by-$n$ matrix, where $d$ is the number of filters and each row of $\ma{W}$ represents the weights of a filter. $\ve{b}$ is a vector of biases, $\ve{y}$ is the response at a pixel of the output map, and $l$ denotes the index of a layer.
$\ve{x}_{l}=f(\ve{y}_{l-1})$
where $f$ is the activation and $c_l = d_{l-1}$. 
As in \cite{heinit,GlorotAISTATS2010}, we assume that initialized elements in $\ma{W}_{l}$ are mutually independent and share the same distribution, as well as that $\ve{x}_l$ are mutually independent and share the same distribution. $\ve{x}_l$ and $\ma{W}_{l}$ are independent of each other. Then: 

\begin{equation}
\operatorname{Var}[y_{l}]=n_l\operatorname{Var}[w_{l}x_l],
\end{equation}
where $y_{l}$, $x_{l}$, and $w_{l}$ represent the random variables of each element in $\ve{y}_l$, $\ma{W}_l$, and $\ve{x}_l$ respectively. Since Kaiming initialization forces $w_{l}$ to have a zero mean, the variance of the product of independent variables gives us:

\begin{equation}\label{eq:y1}
\operatorname{Var}[y_{l}]=n_l\operatorname{Var}[w_{l}]E[x^2_{l}].
\end{equation}
Please note that $E[x^2_{l}]\neq \operatorname{Var}[x_l]$ unless $x_l$ has zero mean. For the ReLU activation (which is used in modern-day architectures), $x_{l}=max(0, y_{l-1})$ and thus may not have zero mean. 
If we let $w_{l-1}$ also have a symmetric distribution around zero and $b_{l-1}=0$, then $y_{l-1}$ has zero mean and has a symmetric distribution around zero. This leads to
$E[x^2_{l}]=\frac{1}{2}\operatorname{Var}[y_{l-1}]$ when $f$ is ReLU.
Combining this with Eqn (\ref{eq:y1}), we obtain:
\begin{equation}\label{eq:y2}
\operatorname{Var}[y_{l}]=\frac{1}{2}n_l\operatorname{Var}[w_{l}]\operatorname{Var}[y_{l-1}].
\end{equation}
With $L$ layers put together, we have:
\begin{equation}\label{eq:prod_fw}
\operatorname{Var}[y_{L}]=\operatorname{Var}[y_{1}]\left(\prod_{l=2}^{L}\frac{1}{2}n_l\operatorname{Var}[w_{l}]\right).
\end{equation}
The above equation is key for Kaiming initialization. A good initialization method should avoid reducing or magnifying the magnitudes of input signals significantly. Hence, for the product in Eqn \ref{eq:prod_fw} to only grow linearly (say by a scalar factor 1), %
a sufficient condition is:
\begin{equation}\label{eq:init_fw}
\frac{1}{2}n_l\operatorname{Var}[w_{l}]=1, \quad \forall l.
\end{equation}

\noindent This leads to a zero-mean Gaussian distribution whose standard deviation 
(std) is $\sqrt{2/{n_l}}$. This is used in Kaiming initialization along with initializing $\ve{b}=0$. Because of this careful initialization, the output variance of a residual block without skip connection, $\operatorname{Var}(\mathcal{F}^{j-1} (\mathbf{x}))$ is approximately equal to the input variance with Kaiming initialization, i.e. 
\begin{equation}
\operatorname{Var}(\mathcal{F}^{j-1}(\mathbf{x})) \approx \operatorname{Var}(\mathbf{x})
\end{equation}

\noindent \textbf{(2) The input and output of the residual block are weakly correlated:} Following \cite{NEURIPS2020_9b861925}, to evaluate the covariance term $\operatorname{Cov}(\mathcal{F}^j(\mathcal{F}^{j-1}(\mathbf{x})), \mathcal{F}^{j-1}(\mathbf{x})))$ in Eqn \ref{eq:6}, we assume that any two coordinates in $\mathbf{x}$ are uncorrelated. With this assumption, one can show that the covariance term is about $O(1/\sqrt{d})$ small. To this end, we consider a single layer: $\mathcal{F(\mathbf{x})} = \ve{W}x$, where elements of $\ve{W}$ are sampled from a Gaussian as: $w_{i j} \sim \mathcal{N}(0,1 / d)$ to match the input/output variance.

\begin{equation}\label{eq:final}
\begin{aligned}
\operatorname{Cov}\left(\left[\mathbf{x}\right]_{i},\left[\mathcal{F}^{j}\left(\mathbf{x}\right)\right]_{i}\right) &= \operatorname{Cov}\left(\left[\mathbf{x}\right]_{i}, \sum_{j=1}^{d} w_{i j}\left[\mathbf{x}\right]_{j}\right) \\ 
&= \sum_{j=1}^{d} w_{i j} \operatorname{Cov}\left(\left[\mathbf{x}\right]_{i},\left[\mathbf{x}\right]_{j}\right) \\ &= w_{i i} \operatorname{Var}\left(\left[\mathbf{x}\right]_{i}\right)
\end{aligned}
\end{equation}

\noindent The above equation holds since we assume any two coordinates in $\mathbf{x}$ are uncorrelated. From Eqn \ref{eq:final}, $\operatorname{Var}(\left[ x_{k+1}\right]_i) \approx (2 + w_{ii})\operatorname{Var}(\left[x_k\right]_i)$. Since $w_{i j} \sim \mathcal{N}(0,1 / d)$, with probability at least $1 - \exp(-d/4)$, $2+w_{ii} > 2 - 1/\sqrt{2}$. In a common ResNet, with $d \geq 32$, the output variance of the residual block increases exponentially with very high probability. If there are multiple layers in the residual block $\mathcal{F}$, ReLU activations would further decrease (at least not increase) the correlation between $\mathbf{x}$ and $\mathcal{F}(\mathbf{x})$. So, the correlations between $\mathbf{x}$ and $\mathcal{F}(\mathbf{x})$ are $O(\sqrt{d})$ small. We request the interested reader to refer to \cite{NEURIPS2020_9b861925} for more details on this analysis.

\noindent The above analysis formally shows the problem of increase in variance when skip connections are introduced. This forms the primary motivation for introducing additional batch-normalization layers in \methodname (as discussed towards the end of Section 2).

\begin{table*}[]
\resizebox{\linewidth}{!}{\begin{tabular}{@{}l|c|c|cc|cc|cc|cc|cc|cc@{}}
\toprule
\textbf{Architecture}           & \textbf{Method}       & \multicolumn{1}{l|}{\textbf{\# Parameters}} & \multicolumn{2}{c|}{\textbf{RESISC-45}}           & \multicolumn{2}{c|}{\textbf{STL-10}}                 & \multicolumn{2}{c|}{\textbf{GTSRB}}                  & \multicolumn{2}{c|}{\textbf{Aircraft}}               & \multicolumn{2}{c|}{\textbf{Flowers}}                & \multicolumn{2}{c}{\textbf{Cars}}                 \\ \midrule
                                & \multicolumn{1}{l|}{} & \multicolumn{1}{l|}{}                       & \multicolumn{1}{c|}{ImageNet}    & RESISC-45      & \multicolumn{1}{c|}{ImageNet}       & STL-10         & \multicolumn{1}{c|}{ImageNet}       & GTSRB          & \multicolumn{1}{c|}{ImageNet}       & Aircraft       & \multicolumn{1}{c|}{ImageNet}       & Flowers        & \multicolumn{1}{c|}{ImageNet}    & Cars           \\ \midrule
\multirow{2}{*}{ResNet-18}      & PAINT                 & 11.2M                                       & \multicolumn{1}{c|}{{\underline{68.87}}} & 14.81          & \multicolumn{1}{c|}{{\underline{68.87}}}    & 48.35          & \multicolumn{1}{c|}{{\underline{68.87}}}    & 10.61          & \multicolumn{1}{c|}{{\underline{68.87}}}    & 1.80           & \multicolumn{1}{c|}{{\underline{68.87}}}    & 6.99           & \multicolumn{1}{c|}{{\underline{68.87}}} & 1.02           \\ \cmidrule(l){2-15} 
                                & \methodname (1)          & 0.4M                                        & \multicolumn{1}{c|}{{\underline{68.87}}} & \textbf{81.10} & \multicolumn{1}{c|}{{\underline{68.87}}}    & \textbf{94.14} & \multicolumn{1}{c|}{{\underline{68.87}}}    & \textbf{71.62} & \multicolumn{1}{c|}{{\underline{68.87}}}    & \textbf{35.16} & \multicolumn{1}{c|}{{\underline{68.87}}}    & \textbf{83.90} & \multicolumn{1}{c|}{{\underline{68.87}}} & \textbf{35.42} \\ \midrule
\multirow{2}{*}{ResNet-34}      & PAINT                 & 21.8M                                       & \multicolumn{1}{c|}{{\underline{74.66}}} & 12.84          & \multicolumn{1}{c|}{{\underline{74.66}}}    & 53.95          & \multicolumn{1}{c|}{{\underline{74.66}}}    & 11.46          & \multicolumn{1}{c|}{{\underline{74.66}}}    & 1.50           & \multicolumn{1}{c|}{{\underline{74.66}}}    & 9.06           & \multicolumn{1}{c|}{{\underline{74.66}}} & 2.26           \\ \cmidrule(l){2-15} 
                                & \methodname (1)          & 0.4M                                        & \multicolumn{1}{c|}{{\underline{74.66}}} & \textbf{81.70} & \multicolumn{1}{c|}{{\underline{74.66}}}    & \textbf{94.49} & \multicolumn{1}{c|}{{\underline{74.66}}}    & \textbf{67.38} & \multicolumn{1}{c|}{{\underline{74.66}}}    & \textbf{32.97} & \multicolumn{1}{c|}{{\underline{74.66}}}    & \textbf{82.34} & \multicolumn{1}{c|}{{\underline{74.66}}} & \textbf{34.62} \\ \midrule
\multirow{2}{*}{ResNet-50}      & PAINT                 & 23.5M                                       & \multicolumn{1}{c|}{{\underline{80.15}}} & 34.30          & \multicolumn{1}{c|}{{\underline{80.15}}}    & 93.46          & \multicolumn{1}{c|}{{\underline{80.15}}}    & 17.05          & \multicolumn{1}{c|}{{\underline{80.15}}}    & 3.84           & \multicolumn{1}{c|}{52.91}          & 61.70          & \multicolumn{1}{c|}{{\underline{80.15}}} & 6.80           \\ \cmidrule(l){2-15} 
                                & \methodname (1)          & 7.1M                                        & \multicolumn{1}{c|}{{\underline{80.15}}} & \textbf{88.29} & \multicolumn{1}{c|}{{\underline{80.15}}}    & \textbf{96.40} & \multicolumn{1}{c|}{{\underline{80.15}}}    & \textbf{75.57} & \multicolumn{1}{c|}{{\underline{80.15}}}    & \textbf{42.81} & \multicolumn{1}{c|}{\textbf{80.15}} & \textbf{85.28} & \multicolumn{1}{c|}{{\underline{80.15}}} & \textbf{47.07} \\ \midrule
\multirow{2}{*}{ResNet-101}     & PAINT                 & 42.5M                                       & \multicolumn{1}{c|}{{\underline{81.87}}} & 27.57          & \multicolumn{1}{c|}{74.66}          & 56.79          & \multicolumn{1}{c|}{78.27}          & 10.26          & \multicolumn{1}{c|}{{\underline{81.87}}}    & 2.54           & \multicolumn{1}{c|}{64.50}          & 52.90          & \multicolumn{1}{c|}{{\underline{81.87}}} & 5.75           \\ \cmidrule(l){2-15} 
                                & \methodname (1)          & 7.1M                                        & \multicolumn{1}{c|}{{\underline{81.87}}} & \textbf{85.13} & \multicolumn{1}{c|}{\textbf{81.87}} & \textbf{97.01} & \multicolumn{1}{c|}{\textbf{81.87}} & \textbf{73.35} & \multicolumn{1}{c|}{{\underline{81.87}}}    & \textbf{39.18} & \multicolumn{1}{c|}{\textbf{81.87}} & \textbf{83.72} & \multicolumn{1}{c|}{{\underline{81.87}}} & \textbf{43.10} \\ \midrule
\multirow{2}{*}{ConvNeXt-Tiny}  & PAINT                 & 27.8M                                       & \multicolumn{1}{c|}{{\underline{82.47}}} & 22.06          & \multicolumn{1}{c|}{\textbf{86.78}} & 82.47          & \multicolumn{1}{c|}{{\underline{82.47}}}    & 6.68           & \multicolumn{1}{c|}{{\underline{82.47}}}    & 3.21           & \multicolumn{1}{c|}{76.69}          & 62.81          & \multicolumn{1}{c|}{{\underline{82.47}}} & 3.58           \\ \cmidrule(l){2-15} 
                                & \methodname (1)          & 4.0M                                        & \multicolumn{1}{c|}{{\underline{82.47}}} & \textbf{91.03} & \multicolumn{1}{c|}{82.47}          & \textbf{98.29} & \multicolumn{1}{c|}{{\underline{82.47}}}    & \textbf{82.48} & \multicolumn{1}{c|}{{\underline{82.47}}}    & \textbf{60.19} & \multicolumn{1}{c|}{\textbf{82.47}} & \textbf{93.15} & \multicolumn{1}{c|}{{\underline{82.47}}} & \textbf{65.49} \\ \midrule
\multirow{2}{*}{ConvNeXt-Small} & PAINT                 & 49.5M                                       & \multicolumn{1}{c|}{{\underline{83.30}}} & 25.95          & \multicolumn{1}{c|}{{\underline{83.30}}}    & 79.71          & \multicolumn{1}{c|}{{\underline{83.30}}}    & 14.93          & \multicolumn{1}{c|}{{\underline{83.30}}}    & 4.20           & \multicolumn{1}{c|}{76.22}          & 63.23          & \multicolumn{1}{c|}{{\underline{83.30}}} & 4.53           \\ \cmidrule(l){2-15} 
                                & \methodname (1)          & 4.0M                                        & \multicolumn{1}{c|}{{\underline{83.30}}} & \textbf{89.11} & \multicolumn{1}{c|}{{\underline{83.30}}}    & \textbf{98.53} & \multicolumn{1}{c|}{{\underline{83.30}}}    & \textbf{80.38} & \multicolumn{1}{c|}{{\underline{83.30}}}    & \textbf{54.40} & \multicolumn{1}{c|}{\textbf{83.30}} & \textbf{91.59} & \multicolumn{1}{c|}{{\underline{83.30}}} & \textbf{60.27} \\ \midrule
\multirow{2}{*}{ConvNeXt-Base}  & PAINT                 & 87.6M                                       & \multicolumn{1}{c|}{{\underline{84.15}}} & 26.52          & \multicolumn{1}{c|}{{\underline{84.15}}}    & 75.54          & \multicolumn{1}{c|}{{\underline{84.15}}}    & 11.25          & \multicolumn{1}{c|}{{\underline{84.15}}}    & 3.39           & \multicolumn{1}{c|}{77.52}          & 80.53          & \multicolumn{1}{c|}{{\underline{84.15}}} & 6.85           \\ \cmidrule(l){2-15} 
                                & \methodname (1)          & 7.1M                                        & \multicolumn{1}{c|}{{\underline{84.15}}} & \textbf{89.43} & \multicolumn{1}{c|}{{\underline{84.15}}}    & \textbf{98.98} & \multicolumn{1}{c|}{{\underline{84.15}}}    & \textbf{80.52} & \multicolumn{1}{c|}{{\underline{84.15}}}    & \textbf{53.23} & \multicolumn{1}{c|}{\textbf{84.15}} & \textbf{90.06} & \multicolumn{1}{c|}{{\underline{84.15}}} & \textbf{62.53} \\ \midrule
\multirow{2}{*}{ConvNeXt-Large} & PAINT                 & 196.2M                                      & \multicolumn{1}{c|}{{\underline{84.52}}} & 22.92          & \multicolumn{1}{c|}{{\underline{84.52}}}    & 77.25          & \multicolumn{1}{c|}{{\underline{84.52}}}    & 9.89           & \multicolumn{1}{c|}{8.02}           & 61.33          & \multicolumn{1}{c|}{80.99}          & 73.87          & \multicolumn{1}{c|}{{\underline{84.52}}} & 5.77           \\ \cmidrule(l){2-15} 
                                & \methodname (1)          & 15.9M                                       & \multicolumn{1}{c|}{{\underline{84.52}}} & \textbf{90.29} & \multicolumn{1}{c|}{{\underline{84.52}}}    & \textbf{98.99} & \multicolumn{1}{c|}{{\underline{84.52}}}    & \textbf{79.40} & \multicolumn{1}{c|}{\textbf{84.52}} & \textbf{52.99} & \multicolumn{1}{c|}{\textbf{84.52}} & \textbf{89.80} & \multicolumn{1}{c|}{{\underline{84.52}}} & \textbf{62.60} \\ \bottomrule
\end{tabular}}
\caption{Top-1 accuracies of the final patched model on the base task (ImageNet) and the patching task. Harmonic mean of these values are presented in Table 2 of main paper. \underline{\quad} denotes the experiments where the accuracy of our proposed method and the baseline (PAINT) are same. \textbf{\methodname significantly outperform PAINT on the patching task, while maintaining the accuracy on the base task.} }
\label{tab:one-task-patching-appendix}
\end{table*}

\section{ Ablation Studies for Patching} \label{sec:patching-abl}
\vspace{-3pt}
\noindent Owing to space constraints, we could not provide a detailed analysis of the patching experiments in the main paper. We present these results in this section.

\subsection{ Single-Task Patching}
\vspace{-3pt}
\noindent Tables 2, 3, 4 in the main paper presents the harmonic mean of the top-1 accuracies of the final patched model on the patching task and the Base Task. While harmonic mean as discussed in the main paper gives equal importance to performance on all the previously learned tasks and the patching task, Top-1 accuracies on the base and patched tasks provide a holistic view. Table \ref{tab:one-task-patching-appendix}  presents top-1 accuracies of the patched model on the base task (ImageNet) and the patching task (dataset for which the model is leveraged for).  In most cases, PAINT with $\alpha=0$ has the highest harmonic mean, where $\alpha=0$ corresponds to the model containing only the weights pertaining to the base task i.e., ImageNet pre-trained weights in this setting. As a result, the ImageNet accuracy is good but the accuracy on the patched task is very low as evident in Table \ref{tab:one-task-patching-appendix}. \methodname perform significantly better than PAINT on the downstream task while also maintaining the performance on the base task. 

\subsection{ Sequential Patching: Understanding the performance drop in PAINT}
\vspace{-3pt}
\noindent In Table 4 of the main paper, we presented results for sequential patching where we report the mean of Top-1 accuracies of the final model on all patching tasks. As PAINT requires choosing an $\alpha$ value empirically, we choose the $\alpha$ value that gives best performance on all the previous tasks (mean based aggregation). To provide better insights into how PAINT performs on all previous tasks with various interpolation coefficient values, we present the top-1 accuracies for all values of $\alpha$ in Table \ref{tab:sequential-patching-all}. As PAINT does not retain the base model weights or weights learned for each task, there is significant drop in performance on all previously learned tasks.

\begin{table*}[]
\captionsetup{font=footnotesize}
\footnotesize
\resizebox{\linewidth}{!}{\begin{tabular}{l|c|cccc|cccc|cccc<{\kern-\tabcolsep}}
\toprule
\textbf{Method (Base Model)}                                                       & \textbf{\begin{tabular}[c]{@{}c@{}}Alpha \\ (interpolation\\ coefficient)\end{tabular}} & \multicolumn{4}{c|}{\textbf{Aircraft-Flowers-STL}}                        & \multicolumn{4}{c|}{\textbf{GTSRB-Cars-Flowers}}                         & \multicolumn{4}{c}{\textbf{STL-Cars-GTSRB}}                            \\ \midrule
\multirow{12}{*}{\begin{tabular}[c]{@{}l@{}}PAINT \\ (ConvNeXT-Tiny)\end{tabular}} & \textbf{}                                                                               & \textbf{ImageNet} & \textbf{Aircraft} & \textbf{Flowers} & \textbf{STL}   & \textbf{ImageNet} & \textbf{GTSRB}  & \textbf{Cars}   & \textbf{Flowers} & \textbf{ImageNet} & \textbf{STL10} & \textbf{Cars}   & \textbf{GTSRB}  \\
                                                                                   & 0.0                                                                                     & 0.09              & 1.26              & 28.59            & 18.31          & 0.11              & 1.91            & \textbf{71.74}           & 13.03            & 0.09              & 11.18          & \textbf{72.96}           & 7.63            \\
                                                                                   & 0.1                                                                                     & 0.10              & 0.99              & 0.33             & 10.00          & 0.10              & 1.98            & 71.62           & 15.53            & 0.07              & 11.34          & 23.06           & 52.37           \\
                                                                                   & 0.2                                                                                     & 0.10              & 0.99              & 0.42             & 10.00          & 0.10              & 1.95            & 71.41           & 18.38            & 0.09              & 10.60          & 2.49            & 89.12           \\
                                                                                   & 0.3                                                                                     & 0.10              & 0.99              & 0.42             & 10.00          & 0.10              & 1.91            & 71.09           & 21.74            & 0.11              & 10.54          & 1.09            & 96.05           \\
                                                                                   & 0.4                                                                                     & 0.10              & 0.99              & 0.42             & 10.00          & 0.09              & 1.77            & 70.40           & 26.26            & 0.13              & 10.53          & 0.80            & 97.34           \\
                                                                                   & 0.5                                                                                     & 0.10              & 0.99              & 0.42             & 10.00          & 0.08              & 1.80            & 69.95           & 30.61            & 0.13              & 10.74          & 0.63            & 97.77           \\
                                                                                   & 0.6                                                                                     & 0.07              & 0.99              & 0.65             & 10.09          & 0.09              & 1.84            & 68.95           & 35.71            & 0.12              & 10.76          & 0.53            & \textbf{97.85}           \\
                                                                                   & 0.7                                                                                     & 0.07              & 1.02              & 1.09             & 12.51          & 0.09              & 1.94            & 67.77           & 40.32            & 0.13              & 10.51          & 0.53            & \textbf{97.85}           \\
                                                                                   & 0.8                                                                                     & 0.08              & 0.54              & 1.68             & 33.81          & 0.11              & 1.92            & 66.57           & 44.30            & 0.12              & 10.36          & 0.53            & 97.71           \\
                                                                                   & 0.9                                                                                     & 0.12              & 1.05              & 2.26             & 65.40          & 0.11              & 2.01            & 65.30           & 47.78            & 0.12              & 10.20          & 0.52            & 97.43           \\
                                                                                   & 1.0                                                                                     & 0.12              & 1.17              & 3.55             & 74.95          & 0.12              & 1.94            & 63.15           & 49.52            & 0.12              & 10.14          & 0.51            & 97.26           \\ \midrule
\rowcolor{highlight}\begin{tabular}[c]{@{}c@{}}ConvNeXT-Tiny \\ +\\\methodname(1)\end{tabular}    & \textbf{-}                                                                              & \textbf{82.47}   & \textbf{60.19}    & \textbf{93.15}   & \textbf{98.29} & \textbf{82.47}   & \textbf{82.48} & 65.49 & \textbf{93.15}   & \textbf{82.47}   & \textbf{98.29} & 65.49 & 82.48 \\ \bottomrule

\multirow{12}{*}{\begin{tabular}[c]{@{}l@{}}PAINT \\ (ResNet-18)\end{tabular}} & \textbf{}                                                                               & \multicolumn{1}{c}{\textbf{}}      & \multicolumn{1}{c}{\textbf{}}      & \multicolumn{1}{c}{\textbf{}}     & \multicolumn{1}{c|}{\textbf{}}      & \textbf{}      & \textbf{}      & \textbf{}      & \textbf{}     & \multicolumn{1}{c}{\textbf{}}      & \multicolumn{1}{c}{\textbf{}}      & \multicolumn{1}{c}{\textbf{}}      & \multicolumn{1}{c}{\textbf{}} \\
                                                                               & 0.0                                                                                     & 0.07                               & 1.38                               & 13.73                             & 37.08                               & 0.06           & 5.70           & 2.54           & 0.55          & 0.08                               & 35.25                              & 1.14                               & 6.13                          \\
                                                                               & 0.1                                                                                     & 0.10                               & 0.99                               & 0.47                              & 10.00                               & 0.09           & 5.69           & 4.24           & 0.55          & 0.10                               & 10.00                              & 0.45                               & 5.46                          \\
                                                                               & 0.2                                                                                     & 0.10                               & 1.00                               & 0.50                              & 10.00                               & 0.09           & 5.73           & 7.72           & 0.55          & 0.10                               & 10.00                              & 0.53                               & 5.46                          \\
                                                                               & 0.3                                                                                     & 0.10                               & 1.00                               & 0.50                              & 10.00                               & 0.07           & 5.66           & 13.98          & 0.57          & 0.10                               & 10.00                              & 0.53                               & 5.46                          \\
                                                                               & 0.4                                                                                     & 0.10                               & 1.00                               & 0.50                              & 10.00                               & 0.07           & 6.32           & 23.70          & 0.85          & 0.10                               & 10.00                              & 0.44                               & 5.46                          \\
                                                                               & 0.5                                                                                     & 0.10                               & 1.00                               & 0.50                              & 10.00                               & 0.08           & 4.62           & 32.48          & 1.72          & 0.10                               & 10.01                              & 0.44                               & 5.46                          \\
                                                                               & 0.6                                                                                     & 0.11                               & 1.00                               & 0.60                              & 10.03                               & 0.07           & 3.20           & 39.25          & 3.43          & 0.10                               & 10.00                              & 0.44                               & 2.85                          \\
                                                                               & 0.7                                                                                     & 0.09                               & 1.01                               & 0.82                              & 11.07                               & 0.09           & 2.15           & \textbf{41.67} & 8.51          & 0.10                               & 10.00                              & 0.44                               & 4.99                          \\
                                                                               & 0.8                                                                                     & 0.10                               & 0.79                               & 1.06                              & 25.95                               & 0.09           & 2.28           & 38.74          & 18.26         & 0.14                               & 12.68                              & 0.55                               & 17.81                         \\
                                                                               & 0.9                                                                                     & 0.12                               & 0.97                               & 1.24                              & 64.39                               & 0.07           & 2.88           & 29.39          & 31.11         & 0.10                               & 10.46                              & 0.36                               & 94.14                         \\
                                                                               & 1.0                                                                                     & 0.12                               & 1.02                               & 1.73                              & 77.20                               & 0.06           & 4.55           & 13.80          & 38.69         & 0.10                               & 8.44                               & 0.47                               & \textbf{98.38}                \\ \midrule
\rowcolor{highlight}\begin{tabular}[c]{@{}c@{}}ResNet-18\\ +\\\methodname(1)\end{tabular}    & \textbf{-}                                                                        & \multicolumn{1}{c}{\textbf{68.87}} & \multicolumn{1}{c}{\textbf{35.16}} & \multicolumn{1}{c}{\textbf{83.9}} & \multicolumn{1}{c|}{\textbf{94.14}} & \textbf{68.87} & \textbf{71.62} & 35.42          & \textbf{83.9} & \multicolumn{1}{c}{\textbf{68.87}} & \multicolumn{1}{c}{\textbf{94.14}} & \multicolumn{1}{c}{\textbf{35.42}} & \multicolumn{1}{c}{71.62}     \\ \bottomrule

\end{tabular}}
\caption{Sequential Patching on ConvNeXT-Tiny and ResNet-18: Top-1 accuracies of the final model for each value of interpolation coefficient ($\alpha$ used by PAINT) on all previously learned tasks, including the base task. \textbf{While PAINT has a significant drop in performance on the previous tasks, \methodname have zero-drop in performance on the previously learned tasks.}}
\label{tab:sequential-patching-all}
\end{table*}

\section{ Training \methodname from Scratch}
\vspace{-3pt}
\noindent All experiments shown so far deal with pre-trained models. It is also important to understand how \methodname behaves when training the model from scratch. To this end, we train and compare standard models and models with \methodname without any form of pre-training. Results for this experiment are shown in Table \ref{tab:scratch-train}.  
\methodname shows consistent improvement even when there is no pre-training involved. Since the fine-grained datasets considered are relatively smaller in size, larger models tend to overfit and perform poorly on test data. \methodname helps in avoiding this overfitting problem when training from scratch with limited data.

\begin{table}
\centering
\begin{tabular}{>{\kern-\tabcolsep}l|cc<{\kern-\tabcolsep}}
\toprule
\textbf{Architecture} & \textbf{FGVC-Aircraft}                & \textbf{Stanford-Cars}                \\ \midrule
ResNet-18                                  & 53.40          & 53.47           \\
\rowcolor{highlight}+ \methodname (4)                            & \textbf{54.12 } & \textbf{55.92 } \\ \midrule
ResNet-34                                  & 48.59           & 43.39           \\
 \rowcolor{highlight} + \methodname (4)                           & \textbf{50.93} & \textbf{54.54} \\ \bottomrule
\end{tabular}
\caption{Comparison of different architectures when trained from scratch without any pre-training.\textbf{\methodname can help when training models from scratch on fine-grained datasets}.}

\label{tab:scratch-train}
\end{table}

\noindent To test the effectiveness of \methodname on commonly used image classification benchmarks, we run experiments on CIFAR-10, CIFAR-100 and ImageNet datasets. 
Results are shown in Table \ref{tab:scratch-acad}. \methodname once again shows improvement on these benchmarks with no pre-training (without significant hyperparameter tuning). (For the Imagenet dataset, we show results only with ResNet-18 architecture due to the significant time and computation requirements otherwise.)

\begin{table}
\resizebox{\linewidth}{!}{\begin{tabular}{>{\kern-\tabcolsep}l|ccc<{\kern-\tabcolsep}}
\toprule
\textbf{Architecture} & \textbf{CIFAR-10} & \textbf{CIFAR-100} & \textbf{ImageNet (Top-5)} \\ \midrule
ResNet-18             & 85.21             & 50.43              & 85.87                     \\
\rowcolor{highlight} + \methodname (4)      & \textbf{85.47}             & \textbf{52.17}              & \textbf{86.16}                     \\ \midrule
ResNet-34             & 81.25             & 48.05              & -                         \\
\rowcolor{highlight} + \methodname (4)      & \textbf{84.76}             & \textbf{51.77}              & -                         \\ \bottomrule
\end{tabular}}
\caption{Comparison of different architectures when trained from scratch without any pre-training on standard image classification datasets.\textbf{\methodname can help improve performance on common image classification benchmark datasets without pre-training}.}

\label{tab:scratch-acad}
\end{table}

\section{ Visualizations} \label{sec:viz}
\vspace{-3pt}
\subsection{ Visualizing Learned $\Lambda$ Values}
\begin{figure}[h]
\captionsetup{font=footnotesize}
  \centering
  \includegraphics[width=0.95\linewidth]{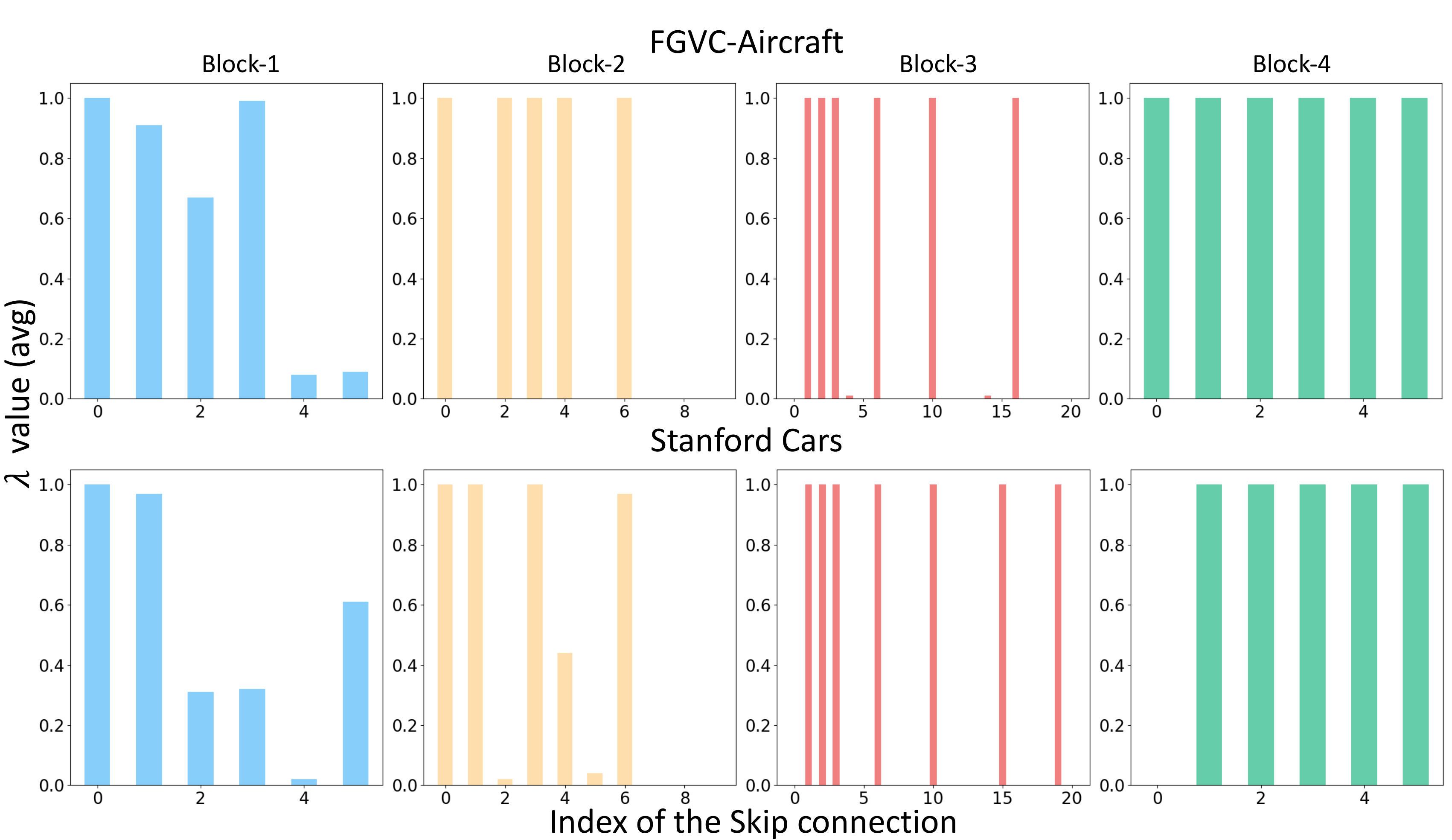}
  \caption{\textbf{Despite not imposing a specific sparsity scheme, \methodname learns to select input-conditioned skip connections in a task-specific manner and zeros out some skip connections (especially in Blocks 2 and 3)}. Qualitative visualization of the weights (average) learned by the \methodname for all test samples of FGVC-Aircraft and Stanford Cars datasets, we use ResNeXt-50 as the base network and introduce four \smethodnames in our \methodname, one for each block}
  \label{fig:vis_lambda}
\end{figure}

\vspace{-3pt}
\noindent To see if \methodname uses all skip connections, we visualize the weights of these skip connections in Fig \ref{fig:vis_lambda}.
We take the average value of weights for all test samples in FGVC-aircraft and Stanford-Dogs datasets.  Even without imposing any specific sparsity schemes such as L1-penalty, some of the skip connection weights are zeros, which shows how the \methodname learns to combine features with few skip connections, which is better than finetuning entire architectures for feature reuse.

\subsection{ Visualization of Learned Design Principles}
\begin{figure*}[h]
  \centering
  \includegraphics[page=1, width=0.9\linewidth]{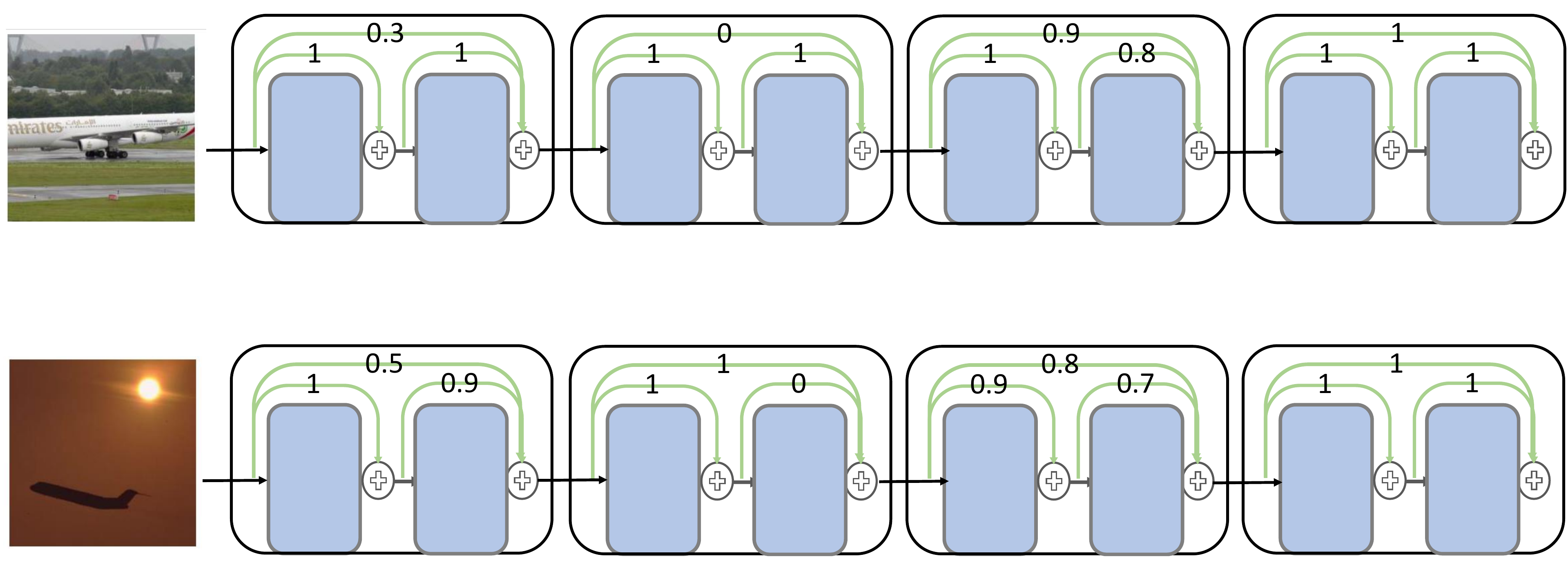} 
  \caption{Visualization of $\Lambda$ values on two samples of FGVC-Aircraft dataset \cite{fgvc-aircraft} for ResNet-18 network}
  \label{figvis1}
\end{figure*}

\begin{figure*}[h]
  \centering
  \includegraphics[page=2, width=0.9\linewidth]{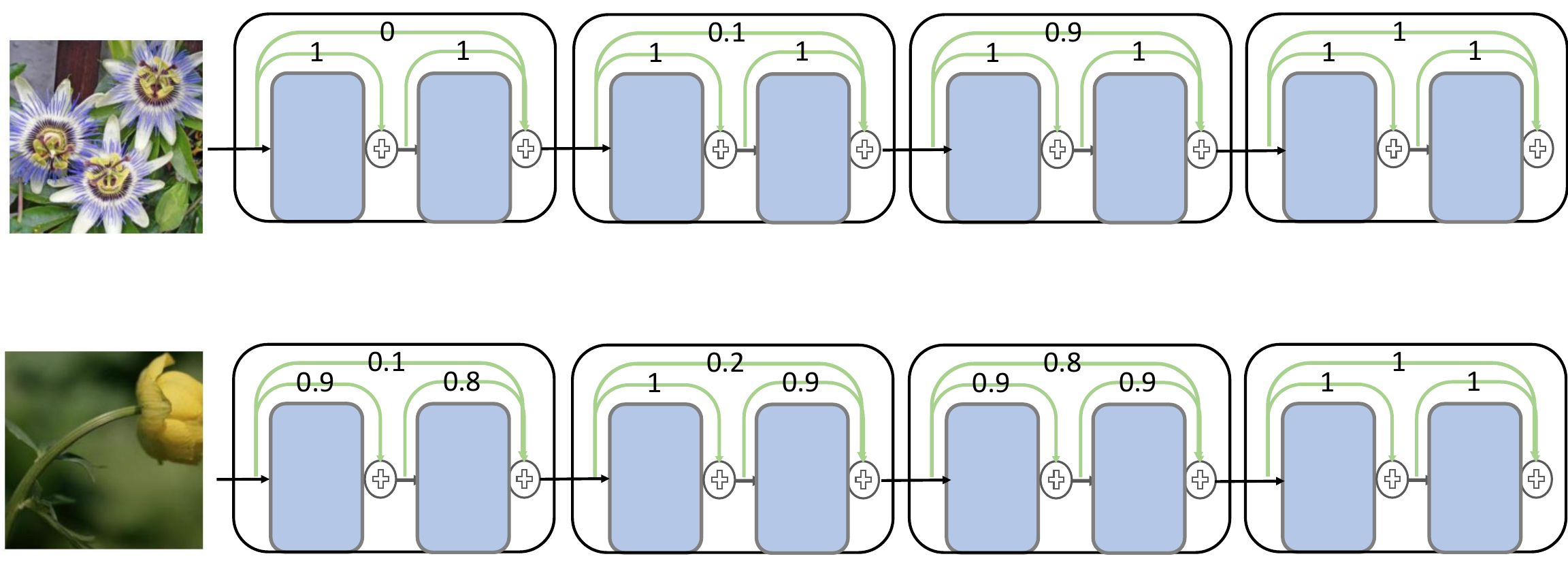} 
    \caption{Visualization of $\Lambda$ values on two samples of Flowers-102 dataset \cite{flowers-102} for ResNet-18 network}
    \label{figvis2}
\end{figure*}

\begin{figure*}[h]
  \centering
  \includegraphics[page=3, width=0.9\linewidth]{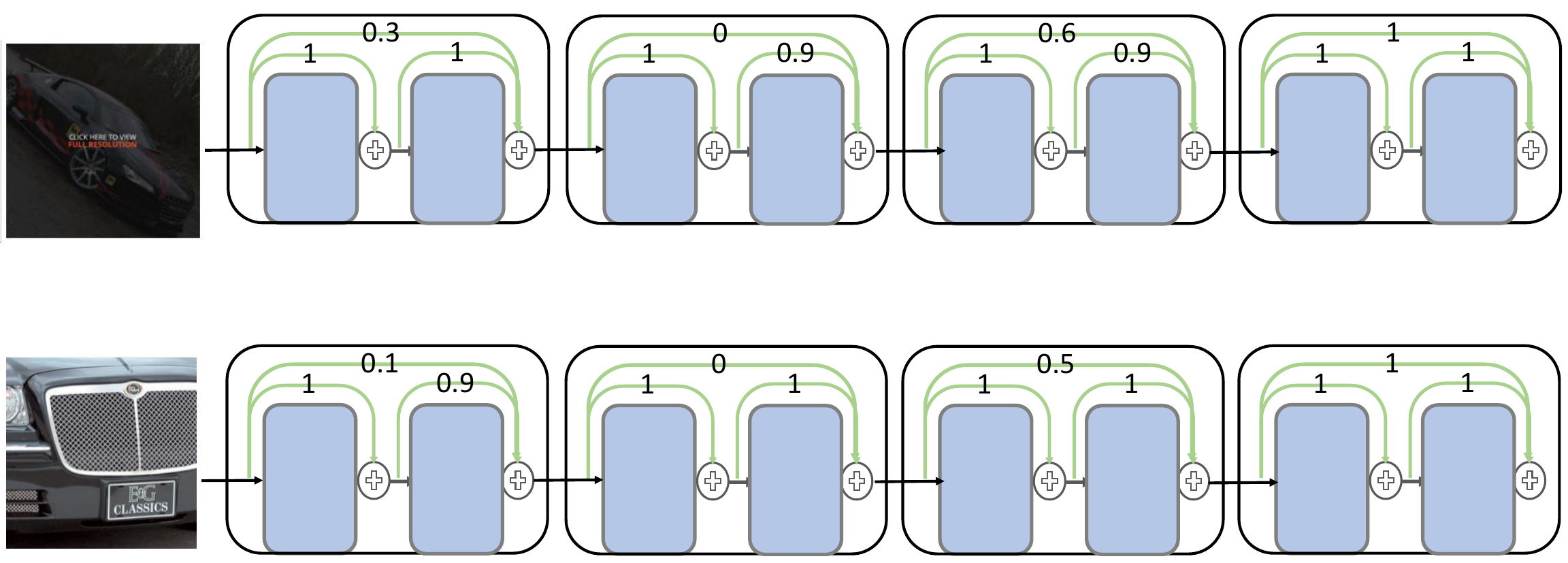} 
    \caption{Visualization of $\Lambda$ values on two samples of Stanford-Cars dataset \cite{stanford-cars} for ResNet-18 network}
    \label{figvis3}
\end{figure*}

\vspace{-6pt}
\noindent We visualize the learned values of $\Lambda$ (weights assigned for the skip connection by the \smethodname) for ResNet-18 on samples from three of the datasets we considered in our experiments.These visualizations are shown in Figures \ref{figvis1}, \ref{figvis2}, \ref{figvis3}. These visualizations provide interesting insights that could be useful for architecture design in general. 
In particular, we note that: (i) In the initial modules, skip connections from the beginning of the block to the end of the block are sufficient, skip connections across blocks have a very less weight; and (ii) In the last module, all skip connections are required with each of them having a weight of 1.
One common trend seen across all datasets is that weights for most skip connections in the last module of the network are always 1 consistently. This can also be observed in Figure \ref{fig:vis_lambda}. 

\newpage

\end{document}